\documentclass{article}
\PassOptionsToPackage{numbers,sort&compress}{natbib}
 \usepackage[preprint]{neurips_2026}

\newcommand{\lith}{\texttt{LithoBench}\xspace}
\usepackage[utf8]{inputenc} 
\usepackage[T1]{fontenc}    
\usepackage{hyperref}       
\usepackage{url}            
\usepackage{booktabs}       
\usepackage{amsfonts}       
\usepackage{nicefrac}       
\usepackage{microtype}      
\usepackage{xcolor}         
\usepackage{graphicx}
\usepackage{amsmath}
\usepackage{graphicx}
\usepackage{newtxtt}
\usepackage{arydshln}
\usepackage{xspace}
\usepackage{enumitem}
\usepackage{float}
\usepackage{wrapfig}
\usepackage{comment}
\usepackage{multirow}

\usepackage[most]{tcolorbox}
\usepackage{listings}
\usepackage{caption}
\usepackage{etoolbox}
\usepackage{etoc}
\usepackage{tocloft}
\definecolor{ceruleanblue}{rgb}{0.16, 0.32, 0.75}
\definecolor{kleinblue}{rgb}{0,0.18,0.65}

\hypersetup{
  colorlinks=true,
  linkcolor=kleinblue,
  citecolor=kleinblue,
  urlcolor=kleinblue
}

\lstdefinestyle{promptstyle}{
  basicstyle=\normalfont\scriptsize,
  breaklines=true,
  breakatwhitespace=false,
  columns=fullflexible,
  keepspaces=true,
  showstringspaces=false,
  tabsize=2,
  frame=none
}

\newtcblisting{promptbox}[2][]{
  enhanced,
  breakable,
  colback=blue!3,
  colframe=blue!70!black,
  boxrule=0.9pt,
  arc=2mm,
  left=2mm,
  right=2mm,
  top=1mm,
  bottom=1mm,
  title={#2},
  fonttitle=\bfseries,
  listing only,
  listing options={
  style=promptstyle,
  breakindent=0pt,
breakautoindent=false,
  },
  #1
}

\usepackage[most]{tcolorbox}

\newtcolorbox{casebox}[2][]{
  enhanced,
  colback=gray!3,
  colframe=gray!60!black,
  boxrule=0.8pt,
  arc=2mm,
  left=2mm,
  right=2mm,
  top=1mm,
  bottom=1mm,
  title={#2},
  fonttitle=\bfseries,
  #1
}
\usepackage{xcolor}
\usepackage{booktabs}
\usepackage{float}

\newcommand{\correctans}[1]{\textcolor{green!45!black}{\textbf{#1}}}
\newcommand{\wrongans}[1]{\textcolor{red!70!black}{\textbf{#1}}}
\newcommand{\modelans}[2]{\textbf{#1}: \quad #2}

\makeatletter
\DeclareRobustCommand\onedot{\futurelet\@let@token\@onedot}
\def\@onedot{\ifx\@let@token.\else.\null\fi\xspace}

\def\eg{\emph{e.g}\onedot}

\makeatother

\title{\lith: Benchmarking Large Multimodal Models for Remote-Sensing Lithology Interpretation}

\author{%
  Jun Wang$^{1}$\thanks{Equal contribution},\,
  Fengpeng Li$^{2}$\footnotemark[1],\,
  Hang Dong$^{3}$,\,
  Tianjin Huang$^{3}$,\,
  Wei Han$^{3}$\thanks{Corresponding author: Wei Han (whan202501@gmail.com)} \\
  $^{1}$School of Computer Science, China University of Geosciences \\
  $^{2}$PRADA Lab, King Abdullah University of Science and Technology \\
  $^{3}$Department of Computer Science, University of Exeter \\
}

\begin{document}

\maketitle

\begin{abstract}

Remote sensing \emph{lithology interpretation} is fundamental to geological surveys, mineral exploration, and regional geological mapping. Unlike general land-cover recognition, lithology interpretation is a knowledge-intensive task that requires experts to infer rock types from various features, \eg, subtle visual,
making reliable automated interpretation highly challenging.
Geological knowledge-guided large multimodal models offer new opportunities, yet their evaluation remains constrained by the lack of benchmarks that capture lithological annotations, multi-level geological semantics, and expert-informed assessment.
Here, we propose \lith, a multi-level benchmark for evaluating geological semantic understanding in remote sensing lithology interpretation.
\lith\ contains 10,000 expert-annotated interpretation instances across 12 representative lithological categories, including \emph{4,000 multiple-choice} and \emph{6,000 open-ended tasks} organized into five cognitive levels: \emph{Identification and Description}, \emph{Comparative Analysis}, \emph{Mechanism Explanation}, \emph{Practical Application}, and \emph{Comprehensive Reasoning}. We further develop an expert-in-the-loop, knowledge-grounded semi-automated construction pipeline, coupling multi sub-processes, \eg, structured geological image descriptions, to enhance geological validity and evaluation reliability.
Experiments with multiple large vision-language models \emph{reveal substantial limitations in geological semantic understanding}, particularly on higher-order explanation, application, and reasoning tasks.
Taken together, \lith\ establishes a dedicated testbed for advancing geological knowledge-guided multimodal models, supporting systematic model evaluation, domain-specific training and adaptation, and knowledge-enhanced reasoning toward expert-level lithology interpretation from remote sensing imagery. 

\end{abstract}

\etocdepthtag.toc{mtchapter}
\section{Introduction}

Remote sensing lithology interpretation is a fundamental component of geological surveys, resource exploration, and regional geological mapping \cite{peyghambari2021hyperspectral, hajaj2024review, chen2023remote}. Its goal goes beyond recognizing surface targets. It requires the joint analysis of texture, tone, structure, geomorphic patterns, and geological context to infer rock types and their genetic implications \cite{ghoneim2024machine}. This task differs from object recognition in natural scenes. Lithological targets are highly domain-specific and often visually subtle. Different rock types may share similar colors and textures. The same rock type may also vary greatly across weathering conditions, terrains, slope aspects, and imaging environments \cite{chen2023remote,ghoneim2024machine}. As a result, lithology analysis requires models to not only perceive visual cues, but also understand their geological meanings. 
Recent large vision-language models (LVLMs) have advanced image understanding, visual question answering, and multimodal reasoning, opening new opportunities for automated lithology interpretation \cite{li2023blip,chen2025vision,liu2023visual,bai2023qwen} (See Sec.~\ref{sec:related_works} of Appendix for related works). Rather than merely predicting rock categories, LVLMs enable more geologically grounded analysis, including feature description, lithological comparison, genetic interpretation, and application-oriented reasoning \cite{zhang2024earthgpt,hu2025rsgpt,zhang2025large,pu2025picabench}.
However, remote sensing lithology interpretation still lacks a dedicated and standardized benchmark for evaluating LVLMs \cite{wang2024earthvqa,li2024vrsbench,han2025novel,hu2025flowsearch}. This gap makes it difficult to assess models across both basic lithological recognition and higher-order geoscientific reasoning, while also limiting systematic comparison, diagnosis, instruction tuning, and task-oriented model improvement.

\begin{figure}[t]
  \centering  \includegraphics[width=0.9\linewidth]{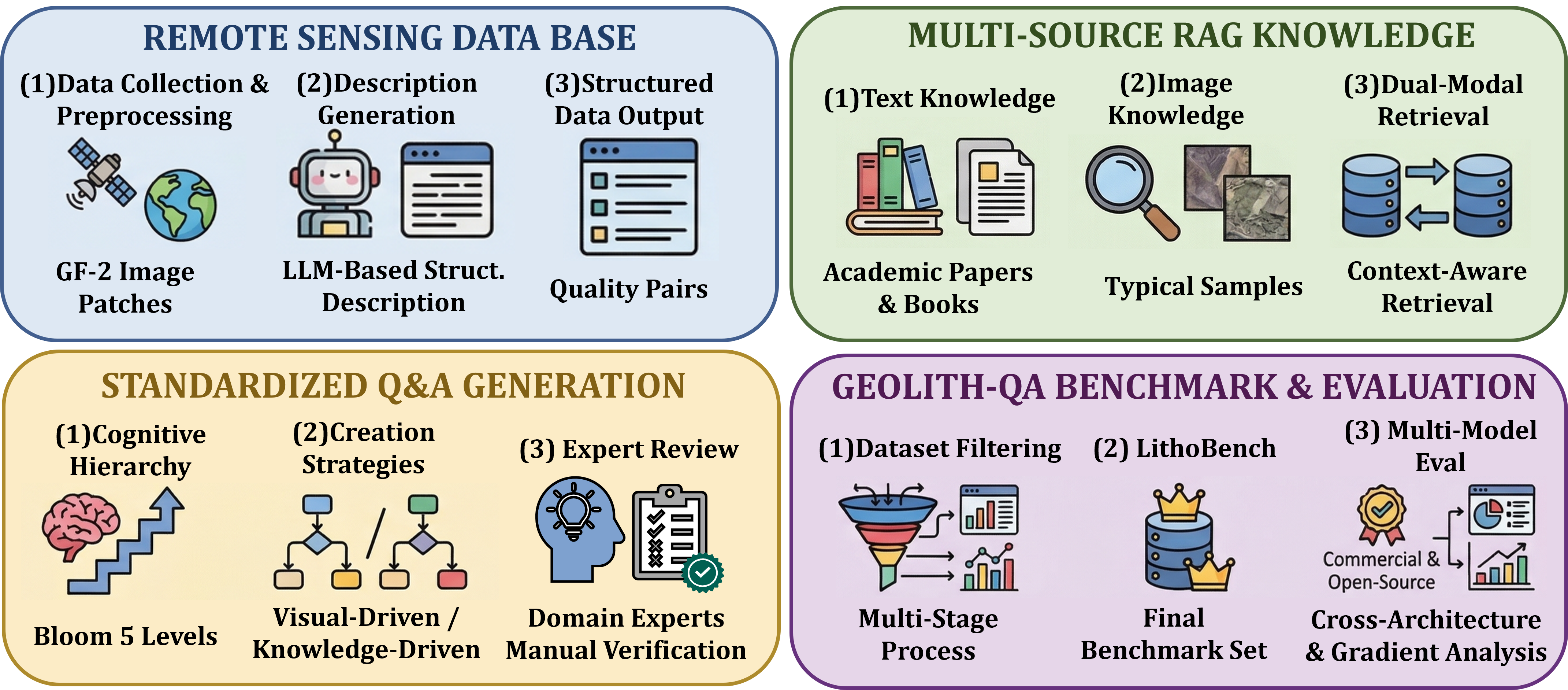}
  \caption{ \lith construction and evaluation pipeline.
  }
  \vspace{-18pt}
  \label{fig:overview}
\end{figure}

Mainstream remote sensing datasets have provided important foundations for perception-oriented tasks such as land-cover classification, object detection, and semantic segmentation, with evaluation protocols typically centered on accuracy, IoU, and related task-specific metrics \cite{xia2023openearthmap,wang2023samrs,sun2022fair1m,lacoste2023geo,marsocci2024pangaea,zhao2025msearth}. However, they are not primarily designed to assess the domain reasoning abilities required for geological interpretation. General multimodal benchmarks provide large-scale evaluation resources \cite{yu2023mm,liu2024mmbench,yue2024mmmu,li2024seed}. Yet their task design and knowledge structure are not well aligned with geological remote sensing, limiting their ability to evaluate professional-level geoscientific understanding. In lithology interpretation, key decisions rarely depend on a single salient object, but instead emerge from the joint analysis of texture, structure, geomorphology, hydrological evidence, and geological context. Therefore, benchmarks relying mainly on generic instruction templates are inadequate for this task. A discriminative and domain-faithful benchmark requires benchmark construction that is tightly grounded in lithological knowledge \cite{wang2024earthvqa,li2024vrsbench,hu2025rsgpt,zhan2025skyeyegpt,zi2025rsvlm,han2023survey}.

\begin{wrapfigure}{r}{0.6\linewidth}
  \centering
  \vspace{-12pt}
  \includegraphics[width=1\linewidth]{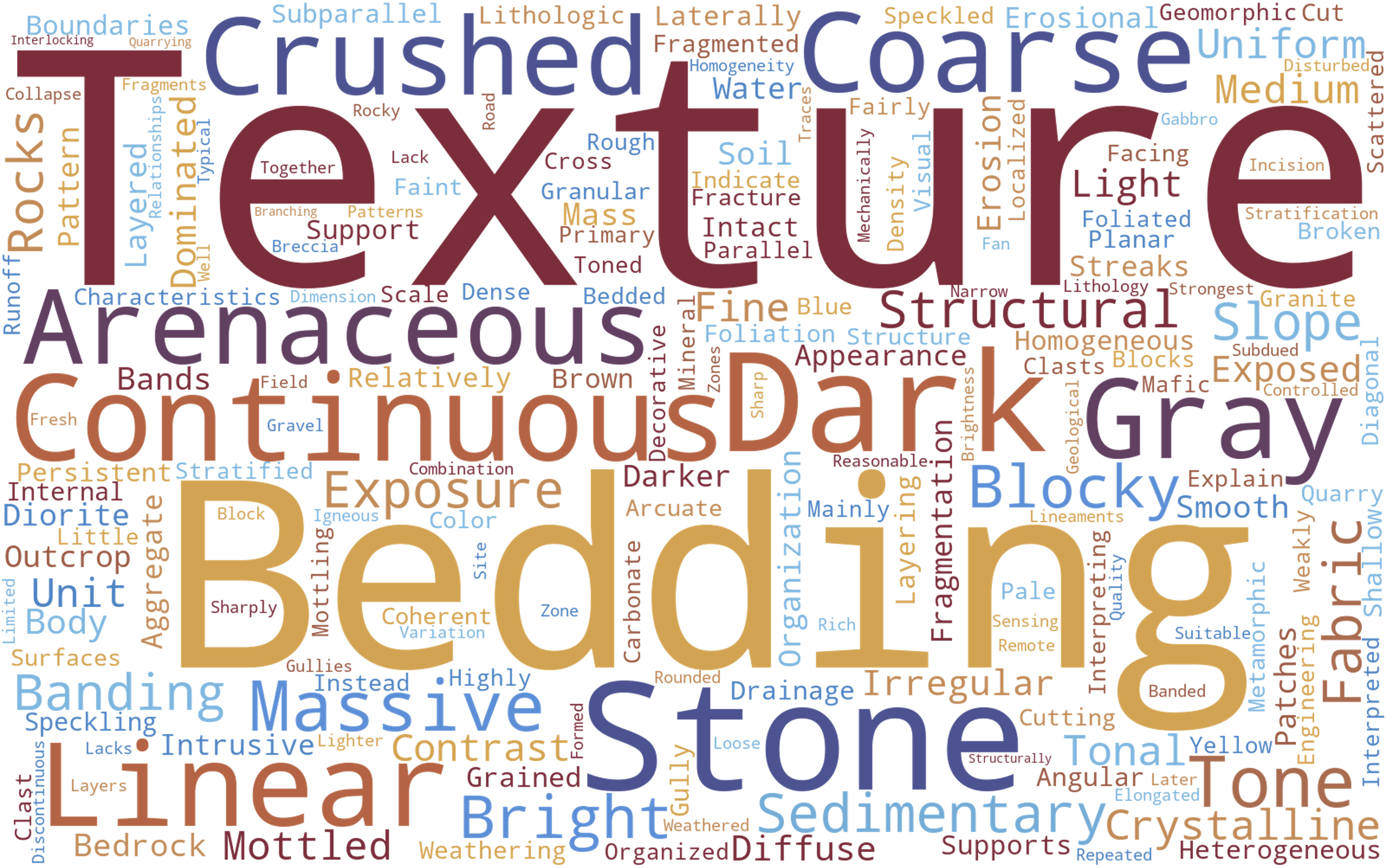}
  \vspace{-12pt}
  \caption{Word cloud of \lith description terms.}
  \label{fig:word_cloud}
  \vspace{-12pt}
\end{wrapfigure}

To address these limitations, we introduce \lith, a large-scale, multi-level benchmark for remote sensing lithology understanding. Unlike recent remote sensing VQA and RS-LVLM benchmarks (e.g., RSVLM‑QA 
\cite{zi2025rsvlm}, VRSBench \cite{li2024vrsbench}, OmniEarth \cite{Fu2026OmniEarth}) that primarily focus on general land-cover or object-level understanding, \lith specifically targets geological lithology interpretation, combining structured visual descriptions, domain knowledge retrieval, and multi-level cognitive tasks to evaluate both visual perception and higher-order geoscientific reasoning.
\lith is built from Gaofen-2 (GF-2) imagery and covers 12 representative lithology and auxiliary land-cover types. It supports two evaluation formats: multiple-choice assessment and open-ended generation, spanning five capability levels: Identification and Description, Comparative Analysis, Mechanism Explanation, Practical Application, and Comprehensive Reasoning. As shown in Figs.~\ref{fig:overview} and ~\ref{fig:word_cloud}, we design a complete data construction pipeline to ensure domain validity and reliable evaluation, consisting of high-quality image sampling, structured visual description, dual-source knowledge retrieval, hierarchical task generation, quality filtering, and model evaluation. By integrating remote sensing evidence with geological knowledge, this design enables standardized assessment of both basic visual recognition and higher-order geoscientific reasoning.
Using \lith, we systematically evaluate a wide range of open-source and closed-source LVLMs and conduct domain fine-tuning experiments \cite{li2023blip,dai2023instructblip,liu2023visual,bai2023qwen,hu2025rsgpt,zhan2025skyeyegpt}. The results reveal clear gaps across capability levels: current general-purpose models remain weak in fine-grained lithological discrimination, genetic interpretation, and application-oriented reasoning. These findings highlight that remote sensing lithology understanding is not a direct extension of general multimodal evaluation, but a domain-specific setting requiring precise visual perception, reliable geological knowledge grounding, and multi-step geoscientific reasoning.

The contributions of this paper are mainly reflected in three aspects:
\begin{itemize}[leftmargin=*]
\item We construct \lith, a multi-level vision-language benchmark for remote sensing lithology understanding, built from a large-scale set of GF-2 image samples and 10,000 high-quality interpretation instances. It provides a unified platform for evaluating LVLMs in visual description, fine-grained discrimination, genetic interpretation, and geoscientific reasoning.

\item We design an expert-in-the-loop, knowledge-grounded semi-automated benchmark construction pipeline that combines structured visual descriptions, geological knowledge retrieval, hierarchical task generation, and expert-guided quality filtering, enabling higher-order lithology reasoning grounded in remote sensing evidence and geological knowledge.

\item We systematically evaluate diverse LVLMs with \lith, revealing key limitations in fine-grained lithological discrimination, genetic interpretation, and geoscientific reasoning, and demonstrating its value for capability diagnosis, domain-specific training, and knowledge enhancement.
\end{itemize}

\section{Pipeline}

\begin{figure}[t]
  \centering
\includegraphics[width=0.9\linewidth]{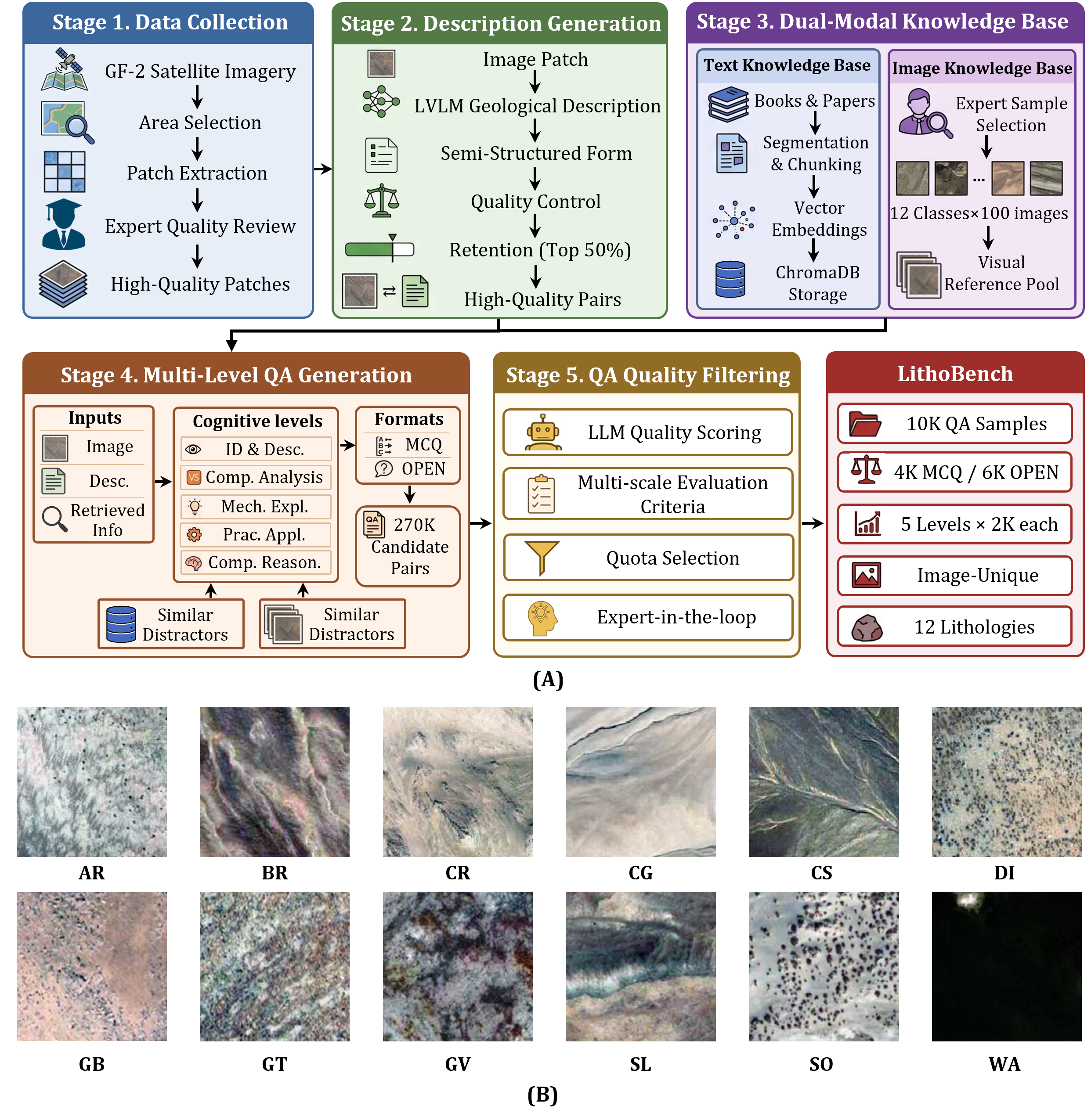}
  \vspace{-8pt}
  \caption{Construction Pipeline and Sample Gallery of \lith.
  (A) Detailed construction pipeline of the \lith.
  (B) Representative images of 12 lithology categories in \lith.}
  \vspace{-14pt}
  \label{fig:pipeline}
\end{figure}

The construction pipeline of \lith is shown in Fig.~\ref{fig:pipeline}(A). It consists of five stages: \emph{image preprocessing, structured description generation, dual-modal retrieval-augmented knowledge base construction, multi-level task generation, and quality filtering and official benchmark construction}. We first crop GF-2 imagery into standardized patches, filter low-quality samples, and organize the remaining samples by category. A multimodal large model is then used to produce structured geological descriptions for each image, which provide stable semantic anchors for data construction. We further build a text knowledge base and an image knowledge base to supply geoscientific context and visually similar reference samples. Guided by five capability levels, namely Identification and Description (ID), Comparative Analysis (Cmp), Mechanism Explanation (Exp), Practical Application (App), and Comprehensive Reasoning (Rea), we generate both multiple-choice and open-ended evaluation tasks. Finally, candidate question-answer pairs are filtered through an expert-in-the-loop quality control process, which combines multi-dimensional LLM-as-a-Judge scoring, expert review, and quota-constrained selection, resulting in the official \lith benchmark.

\subsection{Data Collection and Description Generation}


In this section, as shown in Fig.~\ref{fig:pipeline} (A), we present the details of the Data Collection (\textbf{Stage 1}) first about \emph{Remote Sensing Image Preprocessing}. Then, the Description Generation (\textbf{Stage 2}), consisting of \emph{Description Generator Selection and Description Generation and Quality Control}, is introduced.

\textbf{Data Collection:}~The raw imagery of \lith is collected from GF-2 satellite images over exposed surface regions in northwestern China, covering diverse lithology-related materials and complex geological structures, as shown in Fig.~\ref{fig:pipeline}(B). We define $12$ categories: arenaceous rocks (AR), breccia (BR), carbonate rocks (CR), conglomerate (CG), crushed stone (CS), diorite (DI), gabbro (GB), granite (GT), gravel (GV), slate (SL), soil (SO), and water (WA). To improve label reliability, labels are cross-calibrated using field surveys, core sampling, and remote sensing interpretation, followed by expert inspection to verify patch-label consistency and remove ambiguous cases. The selected regions are cropped into $224 \times 224$ patches, and low-quality samples and edge artifacts are filtered out, resulting in \emph{112,163 high-quality patches} for structured description generation, knowledge retrieval, and QA construction. More details are provided in Appendix~\ref{sec:image_source_preprocessing} and Appendix~\ref{sec:lithology_categories}. With the processed samples, we then perform description generation.

\textbf{Description Generator Selection:}~We compare Gemini-3-Pro \cite{googledeepmind2025gemini3pro} and GPT-5.4 \cite{openai2026gpt54} through expert review and automatic evaluation, using Qwen3-VL-235B-A22B \cite{bai2025qwen3} as the judge model. In 937 blind pairwise comparisons based on professionalism, completeness, and evidence consistency, GPT-5.4 achieves a higher win rate than Gemini-3-Pro (55.92\% vs. 44.08\%). Expert checks on 200 sampled pairs also favor GPT-5.4 in 121 cases. We therefore use GPT-5.4 as the default description generator.

\textbf{Description Generation and Quality Control:}~To obtain structured representations for lithology understanding, we convert each patch into an \emph{expert-guided description}. We use a \emph{semi-structured prompt template}, designed with domain-expert input, to describe tone, texture, pattern, structure, hydrology, geomorphology, and spatial-scale cues in about 500 characters. Each output follows a unified \textbf{\emph{``field--enumerated value--evidence phrase''}} format, balancing natural-language flexibility with automatic parsing and retrieval. Following this protocol, we generate descriptions for over 110,000 image patches. We then re-evaluate each image-description pair with GPT-5.4, which assigns an image quality score $S_{iq}$ and a description quality score $S_{dq}$ to compute the final score $S$ as:
\begin{equation}
S = \alpha \times S_{iq} + (1-\alpha) \times S_{dq} + \delta
\end{equation}
where $\delta$ is set to $+0.3$, $0$, and $-0.5$ for \textbf{\emph{``keep''}, \emph{``manual review''}}, or \textbf{\emph{``discard''}}, respectively. We set $\alpha=0.6$ to emphasize visual evidence quality while retaining description completeness. Finally, we rank all image-description pairs by $S$ and keep the top 50\%, yielding about \emph{56,000 high-quality pairs} for geoscientific understanding instance construction.

\subsection{Dual-Modal Knowledge Base}

As shown in Fig.~\ref{fig:pipeline}(A), we introduce the construction of the Dual-Modal Knowledge Base (\textbf{Stage 3}). For explanation, application, and reasoning tasks, visible image cues are often insufficient, and external geoscientific context is needed. We therefore build two complementary retrieval sources~\cite{lewis2020retrieval}: a \emph{Text Knowledge Base} and an \emph{Image Knowledge Base}. The former provides geoscientific context for explanation, application, and reasoning tasks, while the latter is built from expert-curated samples to support similar-sample retrieval, fine-grained comparison, and realistic distractor generation.

\textbf{Text Knowledge base:}~The text knowledge base includes approximately 1,000 papers on petrology and remote sensing geology published within the past five years, together with more than 20 professional books on rock spectroscopy. We extract and chunk the documents, encode all chunks with BGE-Large-En-v1.5~\cite{xiao2024c}. During instance construction, the query combines the category label, structured description, and target capability level to retrieve the Top-3 chunks as geological context.

\textbf{Image knowledge base:}~The image knowledge base is expert-curated and contains 1,200 representative samples, with 100 high-quality images for each of the 12 categories. All reference images are encoded with CLIP~\cite{radford2021learning}, and retrieval is performed by cosine similarity. Given a target image, the system retrieves the Top-3 visually similar samples from different categories. These references support fine-grained discrimination, realistic distractor construction, and Comparative Analysis.
\subsection{Multi-Level QA Generation}

Based on the revised Bloom's taxonomy~\cite{anderson2001taxonomy}, we design the Multi-Level QA Generation stage (\textbf{Stage 4} of Fig.~\ref{fig:pipeline}(A)) . Domain experts adapt these five capability levels to lithology interpretation by defining capability-specific question templates and expected reasoning focuses. This hierarchy covers a progressive understanding process, from recognizing visible lithological cues, to explaining their geological causes, and to assessing practical implications. It enables both overall model comparison and fine-grained diagnosis of failures in visual perception, knowledge grounding, and reasoning.

For Identification and Description, and Comparative Analysis, instance construction mainly uses visual evidence. For multiple-choice evaluation, the system retrieves visually similar samples from different categories and combines them with the target image into a $2 \times 2$ composite image. These real distractors make the options more diagnostic than random choices. For open-ended evaluation, the system uses the target image and its structured description. Comparative analysis further adds the \textbf{\emph{Top-1}} similar sample to highlight key visual and geological differences.


For Mechanism Explanation, Practical Application, and Comprehensive Reasoning, the system uses text-based retrieval. It retrieves the \emph{Top-3 knowledge fragments} based on the image label, structured description, and target capability level. These fragments are combined with the image and template constraints to ground each instance in geoscientific evidence. This reduces unconstrained generation and makes higher-order instances more traceable.


\subsection{QA Quality Filtering and Official Benchmark Construction}
\begin{figure}[t]
 \centering
 \includegraphics[width=0.9\linewidth]{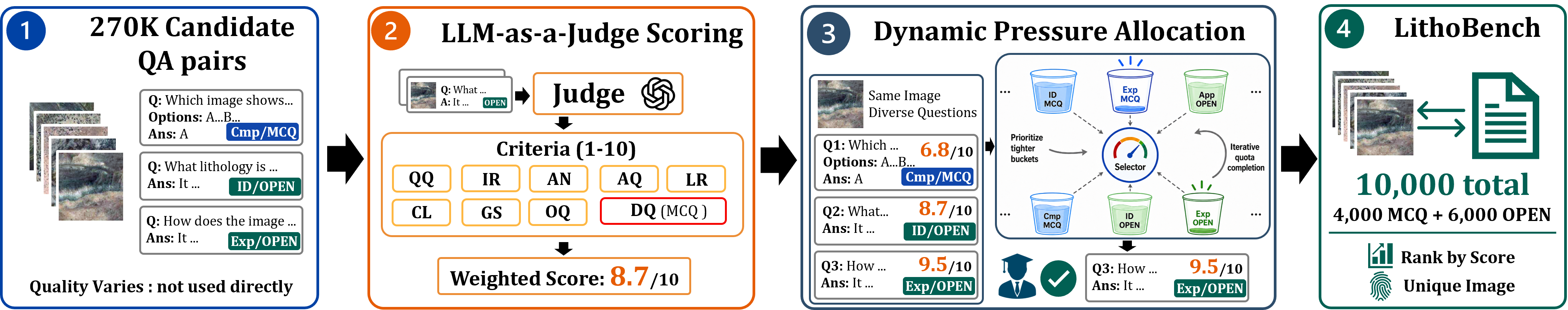}
 \caption{Quality Filtering and Official Benchmark Construction of \lith.}
   \vspace{-18pt}
 \label{fig:quality_filtering}
\end{figure}

Following the analysis in Fig.~\ref{fig:quality_filtering}, the automated generation stage yields about 270,000 candidate instances. These candidates vary in quality, image alignment, answer stability, and analytical rigor. They are therefore not directly used as the final benchmark. We apply an expert-in-the-loop filtering process that combines LLM-as-a-Judge quality scoring, expert review, and quota-constrained selection~\cite{liu2023g,kim2023prometheus}. After automatic scoring, domain experts review representative high-scoring and borderline candidates, focusing on geological correctness, image-question alignment, answer reliability, and potential leakage risk. The expert feedback is used to refine filtering rules and remove candidates with ambiguous lithological evidence, unstable answers, or weak geological grounding.Details of the expert panel, review protocol, and validation statistics are provided in Appendix~\ref{sec:expert_panel}.

Each candidate is scored along multiple dimensions: question quality (QQ), image relevance (IR), clarity (CL), geological soundness (GS), answerability (AN), analysis quality (AQ), overall quality (OQ), and benchmark leak risk (LR). Distractor quality (DQ) is additionally used for multiple-choice instances. All dimensions are scored from 1 to 10. Since a higher LR means greater leakage risk, we define a low-leakage bonus as $LLB = 11 - LR$. Then, the final score for each instance is as:
\begin{equation} 
\begin{aligned} 
\mathrm{WeightedScore}_{\mathrm{OPEN}} =&\ 0.26 \cdot OQ + 0.22 \cdot QQ + 0.18 \cdot IR + 0.08 \cdot CL \\ &+ 0.08 \cdot GS + 0.08 \cdot AN + 0.07 \cdot AQ + 0.03 \cdot LLB . 
\end{aligned} 
\end{equation}
The weighting scheme prioritizes semantic quality and visual grounding. OQ, QQ, and IR receive the largest weights because they reflect overall suitability, formulation quality, and image correspondence. CL, GS, AN, and AQ further ensure clarity, geological validity, answerability, and analytical support. The $LLB$ term provides a leakage-risk adjustment. For multiple-choice instances, DQ is also included to favor plausible and non-trivial distractors.

After scoring, we first select 15,000 high-scoring candidates through local deduplication within each capability-level and instance-type bucket. Domain experts then review these candidates based on both quality scores and professional experience, yielding the final benchmark of 10,000 instances, including \emph{4,000 multiple-choice instances} and \emph{6,000 open-ended instances}. Each of the five capability levels contains 800 multiple-choice and 1,200 open-ended instances. We also enforce global image uniqueness to avoid repeated images across the benchmark. A dynamic quota allocation strategy is used to fill under-resourced buckets while balancing instance quality, category coverage, format proportion, and image diversity.



\section{Dataset Statistics}

\begin{wraptable}{r}{0.4\textwidth}
\centering
\vspace{-12pt}
\caption{Statistics of \lith.}\vspace{2pt}
\label{tab:lith_statistics}
\small
\begin{tabular}{c c}
\toprule
\textbf{Item} & \textbf{Number} \\
\midrule
Categories & 12 \\
Image patches & 112,163 \\
Image-description pairs & $\sim$56,000 \\
Candidate instances & 269,790 \\
Final instances & 10,000 \\
MCQ instances & 4,000 \\
Open-ended instances & 6,000 \\
Capability levels & 5 \\
Instances per level & 2,000 \\
MCQ per level & 800 \\
Open-ended per level & 1,200 \\
\bottomrule
\end{tabular}
\vspace{-12pt}
\end{wraptable}

Table~\ref{tab:lith_statistics} summarizes the scale of \lith. The benchmark covers 12 representative lithology and land-cover types. It starts from 112,163 high-quality remote sensing patches. After description generation and quality filtering, about 56,000 image-description pairs are retained. These pairs produce 269,790 candidate instances. Multi-stage filtering further selects 10,000 final instances, including 4,000 multiple-choice and 6,000 open-ended instances.

\begin{figure}[t]
  \centering
  \includegraphics[width=0.9\linewidth]{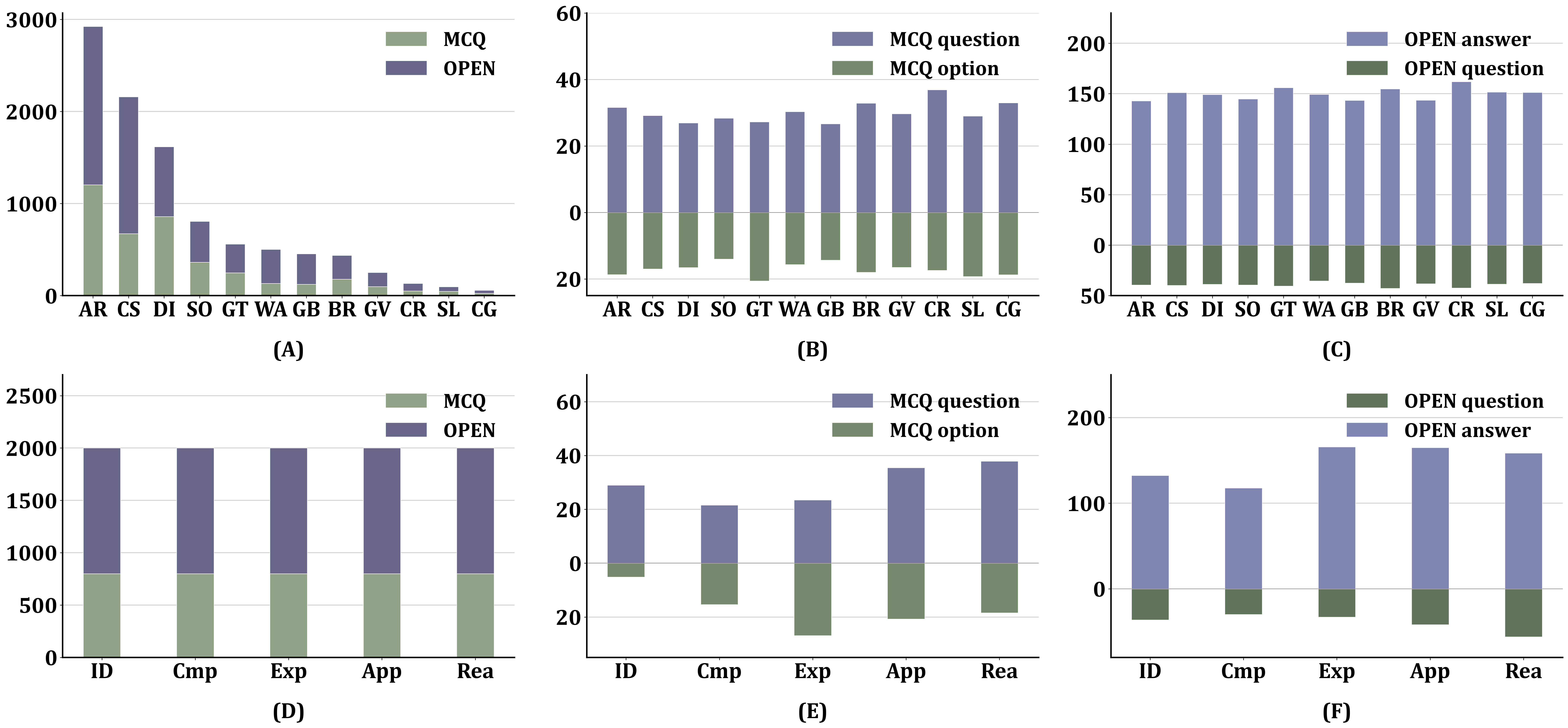}
\vspace{-4pt}
  \caption{Distribution and word-count statistics of \lith samples. 
(A) Distribution of QA samples by lithology category. 
(B) Average word counts of MCQ questions and options by lithology type. 
(C) Average word counts of OPEN questions and answers by lithology type. 
(D) Distribution of samples by question type. 
(E) Average word counts of MCQ questions and options by question category. 
(F) Average word counts of OPEN questions and answers by question category.}
\vspace{-16pt}
  \label{fig:LithoBench_distribution}
\end{figure}

As shown in Fig.~\ref{fig:LithoBench_distribution}(A--C), the instance distribution differs across categories. Arenaceous rocks, crushed stone, and diorite contain relatively more samples. The figure also shows the average word counts of different instance formats. Multiple-choice prompts and options are compact. Open-ended responses are longer because they require more detailed geological interpretation. Fig.~\ref{fig:LithoBench_distribution}(D--F) shows that \lith is balanced across five capability levels. Each level contains 2,000 instances, with 800 multiple-choice and 1,200 open-ended instances. Higher-level instances, especially application and reasoning, tend to involve longer prompts and more detailed responses.

\section{Benchmark Performances}
\label{sec:bench_perform}
\subsection{Experimental Setup}
\label{sec:experiement_setup}
To validate the effectiveness of the \lith benchmark, we systematically evaluate multiple closed-source and open-source large vision-language models under a unified inference setting. The evaluated models include Claude-Sonnet-4.6\cite{anthropic2026sonnet46}, Gemini-3-Pro \cite{googledeepmind2025gemini3pro} , GPT-5.4\cite{openai2026gpt54} , Grok-4 \cite{xai2025grok4}, Qwen3.5-27B \cite{bai2025qwen3}, Qwen3.5-397B-A17B\cite{bai2025qwen3}, Intern-S1-Pro \cite{zou2026intern}, InternVL3.5-241B-A28B\cite{wang2025internvl35}, Kimi-K2.5\cite{moonshot2026kimik25}, and GLM-4.6V\cite{zai2025glm46v}. 
In addition, we conduct LoRA fine-tuning on Qwen3.5-9B \cite{bai2025qwen3} and InternVL3.5-8B-HF \cite{wang2025internvl35} using the \lith training data, with each model trained for $3$ epochs, to examine the effect of domain-specific supervision on improving remote sensing lithology interpretation ability \cite{hu2022lora}. More information can be found in Sec.~\ref{sec:ex_settings}.

\subsection{Evaluation Metrics}
\label{sec:experiment_metrics}
\lith contains both multiple-choice questions (MCQ) and open-ended questions (OPEN), so we use question-type-specific evaluation metrics. For OPEN questions, we report two scores. The first uses reference-based automatic metrics without LLM evaluation:
\begin{equation} 
\mathrm{OPEN}_{\mathrm{w/oJ}} = 0.6 \times \mathrm{ROUGE\text{-}L} \times 10 + 0.4 \times \mathrm{BLEU\text{-}4} \times 10 . 
\end{equation}
The second incorporates LLM-as-a-Judge:
\begin{equation} 
\mathrm{OPEN}_{\mathrm{J}} = 0.4 \times \mathrm{ROUGE\text{-}L} \times 10 + 0.2 \times \mathrm{BLEU\text{-}4} \times 10 + 0.4 \times \mathrm{JudgeOverallScore}. 
\end{equation}
BLEU-4 (B-4) measures phrase-level $n$-gram matching, while ROUGE-L (R-L) measures longest-common-subsequence overlap and better captures global structure consistency. Details are in Sec.~\ref{sec:d_eva_metrics}.

\subsection{Result Analysis}
\label{sec:results_analysis}

\begin{figure}[t]
  \centering
  \includegraphics[width=0.9\linewidth]{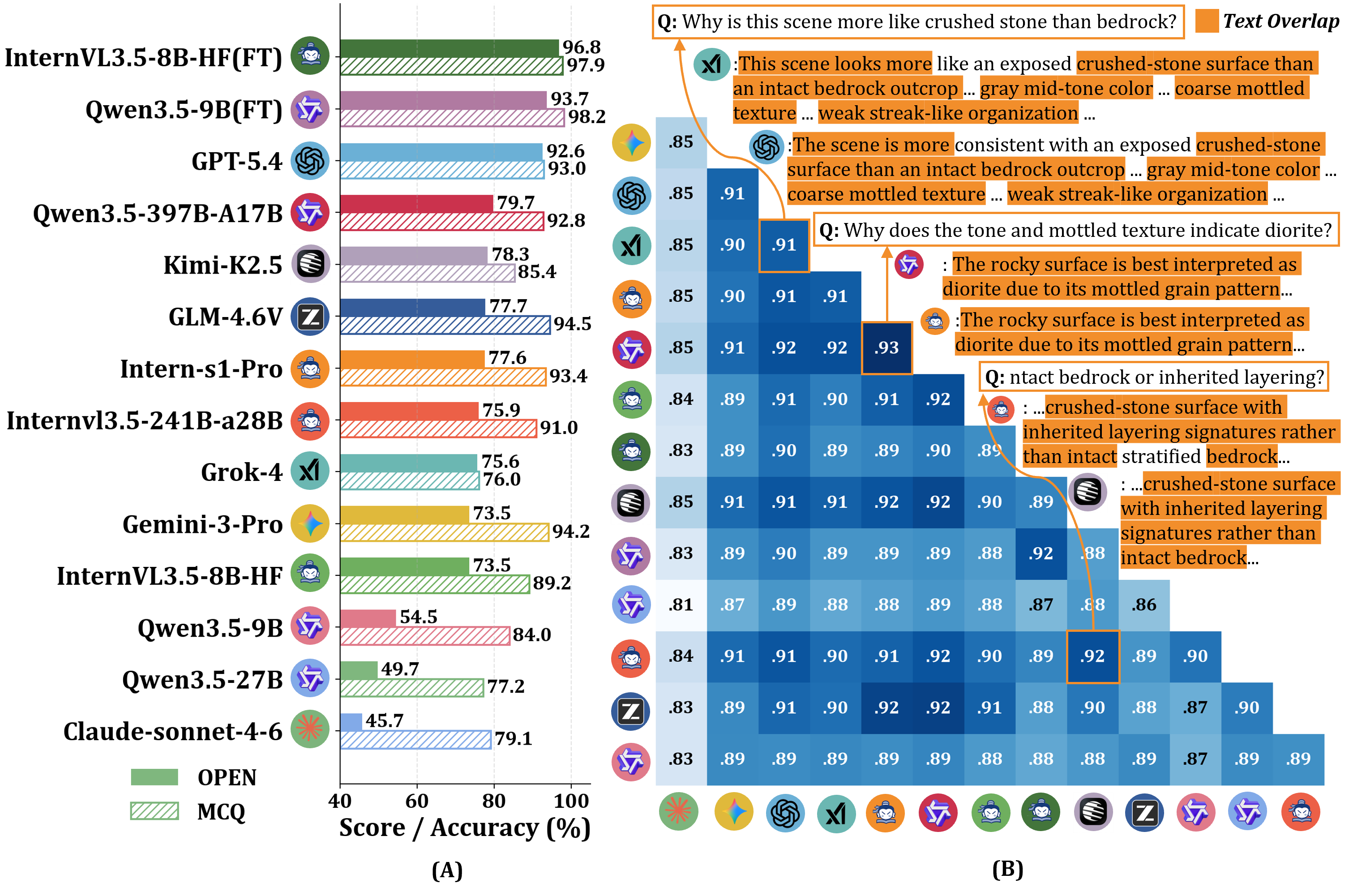}
  \caption{
  Overall performance comparison and response similarity of VLMs on \lith.
(A) Overall OPEN$_{\mathrm{J}}$ and MCQ$_{\mathrm{all}}$ performance.
(B) Pairwise similarity of open-ended responses with representative text-overlap examples.
  }
\vspace{-10pt}
\label{fig:model_performance}
\end{figure}

As shown in Fig.~\ref{fig:model_performance}(A) and Table~\ref{tab:geolithqa_results}, \lith provides clear discrimination across models. For \textbf{MCQ evaluation}, small models fine-tuned with LoRA on the \lith training set outperform unfine-tuned open-source and closed-source models. This suggests that \textbf{domain-specific supervision} improves the use of lithology labels, visual features, and diagnostic cues. The gains are more evident in visually grounded levels, such as identification and description and Comparative Analysis, where objective accuracy is directly tied to fine-grained visual discrimination.

For \textbf{OPEN evaluation}, automatic metrics and LLM-as-a-Judge scores show broadly consistent trends. However, JudgeOverallScore better captures differences in professional expression and reasoning structure. Some models obtain reasonable ROUGE-L or BLEU-4 scores but lower JudgeOverallScore. This indicates that lexical overlap alone does not guarantee \textit{complete}, \textit{geologically sound}, and \textit{logically coherent} explanations. In contrast, stronger models may not always achieve the best automatic scores, but receive higher LLM-based judgments. Their outputs better match expected geological semantics, terminology usage, and reasoning organization.

The capability-level results in Table~\ref{tab:geolithqa_results} reveal a clear trend. Models obtain higher scores on Identification and Description, Comparative Analysis. Their scores decrease on Mechanism Explanation, Practical Application, and Comprehensive Reasoning. This indicates that current LVLMs can capture basic lithological cues in remote sensing images. However, they remain limited when tasks require geological knowledge grounding, genetic interpretation, or multi-step geoscientific reasoning.

We further analyze \emph{cross-model answer similarity} on open-ended instances. As shown in Fig.~\ref{fig:model_performance}(B), we encode each model response with BGE-large-en-v1.5. Responses are aligned by shared instance IDs. We then compute cosine similarity between normalized embeddings and average the results over all shared instances. This gives the pairwise similarity matrix on the right. Most model pairs show similarities between $0.83$ and $0.93$. This indicates substantial overlap in their response semantics. The overlap is mainly related to common visual evidence, such as color, texture, bedding, fragmentation, and erosional morphology. The highlighted examples show that different models may reach similar category-level judgments and use similar geological cues in their explanations. This analysis provides an additional view of model behavior. It helps examine whether open-ended outputs reflect fine-grained geological reasoning or follow common descriptive patterns.

\begin{table*}[t]
  \caption{Performance comparison of different VLMs on \lith.}
  \vspace{-1pt}
  \label{tab:geolithqa_results}
  \centering
  \resizebox{\textwidth}{!}{
  \begin{tabular}{lccccccccccccccc}
    \toprule
    Model 
    & ID-MCQ & ID-OPEN 
    & Cmp-MCQ & Cmp-OPEN 
    & Exp-MCQ & Exp-OPEN 
    & App-MCQ & App-OPEN 
    & Rea-MCQ & Rea-OPEN 
    & R-L & B-4 
    & OPEN$_{\mathrm{w/oJ}}$ 
    & OPEN$_{\mathrm{J}}$ 
    & MCQ$_{\mathrm{all}}$ \\
    \midrule
    Claude-Sonnet-4.6 
    & 0.5625 & 1.9709 & 0.9500 & 1.2307 & 0.6687 & 2.2553 & 0.8250 & 2.8160 & 0.9513 & 3.1611 & 0.1360 & 0.0331 & 0.9484 & 2.2868 & 0.7915 \\

    Gemini-3-Pro 
    & 0.7762 & 3.0566 & 0.9900 & 3.5803 & 0.9775 & 4.0867 & 0.9875 & 3.4877 & 0.9775 & 4.1679 & 0.1644 & 0.0427 & 1.1570 & 3.6758 & 0.9417 \\

    GPT-5.4 
    & 0.6650 & 4.1901 & \textbf{1.0000} & 4.4384 & \textbf{0.9875} & 4.7674 & \textbf{1.0000} & 4.6048 & 0.9975 & \textbf{5.1576} & 0.2231 & 0.0782 & 1.6512 & 4.6317 & 0.9300 \\

    Grok-4 
    & 0.6700 & 3.1364 & 0.8087 & 3.5818 & 0.7050 & 4.0939 & 0.8375 & 3.6916 & 0.7812 & 4.4050 & 0.1635 & 0.0439 & 1.1564 & 3.7817 & 0.7605 \\

    Qwen3.5-27B 
    & 0.4313 & 2.7385 & 0.9513 & 2.4256 & 0.8025 & 3.3867 & 0.9337 & 1.8366 & 0.7412 & 2.0429 & 0.1246 & 0.0298 & 0.8666 & 2.4861 & 0.7720 \\

    Qwen3.5-397B-A17B 
    & 0.6538 & 3.2891 & 0.9988 & 3.8664 & \textbf{0.9875} & 4.2763 & \textbf{1.0000} & 3.8569 & \textbf{0.9988} & 4.6397 & 0.1872 & 0.0545 & 1.3413 & 3.9857 & 0.9277 \\

    Intern-S1-Pro 
    & 0.7038 & 3.2338 & 0.9838 & 3.7302 & \textbf{0.9875} & 4.0997 & 0.9975 & 3.9070 & 0.9963 & 4.4220 & 0.1706 & 0.0471 & 1.2119 & 3.8786 & 0.9338 \\

    InternVL3.5-241B-A28B 
    & 0.6288 & 2.9719 & 0.9788 & 3.6844 & \textbf{0.9875} & 4.1166 & 0.9888 & 3.9562 & 0.9650 & 4.2565 & 0.1670 & 0.0441 & 1.1784 & 3.7971 & 0.9098 \\

    Kimi-K2.5 
    & 0.7588 & 3.0612 & 0.9213 & 3.7907 & 0.8038 & 4.2371 & 0.9063 & 3.9526 & 0.8775 & 4.5452 & 0.1775 & 0.0471 & 1.2536 & 3.9174 & 0.8535 \\

    GLM-4.6V 
    & 0.7625 & 3.2616 & 0.9788 & 3.7512 & \textbf{0.9875} & 4.1096 & 0.9988 & 3.8603 & 0.9975 & 4.4513 & 0.1697 & 0.0443 & 1.1953 & 3.8868 & 0.9450 \\
    
    InternVL3.5-8B-HF
    & 0.5813 & 3.0962 & 0.9950 & 3.7948 & 0.9863 & 3.9614 & 0.9688 & 3.2989 & 0.9275 & 4.2231 & 0.1838 & 0.0425 & 1.2727 & 3.6749 & 0.8918 \\
    
    Qwen3.5-9B
    & 0.9250 & 2.7258 & 0.9075 & 3.3125 & 0.7663 & 3.5929 & 0.7425 & 2.8763 & 0.8575 & 3.6242 & 0.1414 & 0.0297 & 0.9671 & 3.2263 & 0.8398 \\
    
    InternVL3.5-8B-HF(FT)
    & \textbf{0.9863} & \textbf{4.8959} & \textbf{1.0000} & 4.6128 & 0.9400 & \textbf{4.9254} & 0.9775 & \textbf{4.6670} & 0.9888 & 5.1100 & \textbf{0.2570} & \textbf{0.1080} & \textbf{1.9740} & \textbf{4.8422} & 0.9785 \\

    Qwen3.5-9B(FT) 
    & 0.9812 & 4.4109 & 0.9988 & \textbf{4.6552} & 0.9712 & 4.7055 & 0.9800 & 4.5644 & 0.9812 & 5.0936 & 0.2306 & 0.0863 & 1.7288 & 4.6859 & \textbf{0.9825} \\
    \bottomrule
  \end{tabular}
  }
  \vspace{-5pt}
\end{table*}

Overall, \lith contributes in three aspects. First, it evaluates both \textbf{objective discrimination} and \textbf{open-ended professional understanding}, rather than only single-label prediction. Second, training and benchmark images are strictly separated. Thus, the gains from domain fine-tuning better reflect task adaptation rather than image memorization. Third, the combination of \textbf{five capability levels}, \textbf{two evaluation formats}, and \textbf{multiple metrics} reveals model strengths and limitations across recognition, explanation, and reasoning. This provides a structured basis for developing domain-specific multimodal models for remote sensing.

\subsection{Further Discussion}

\begin{figure}[t]
  \centering
  \includegraphics[width=1\linewidth]{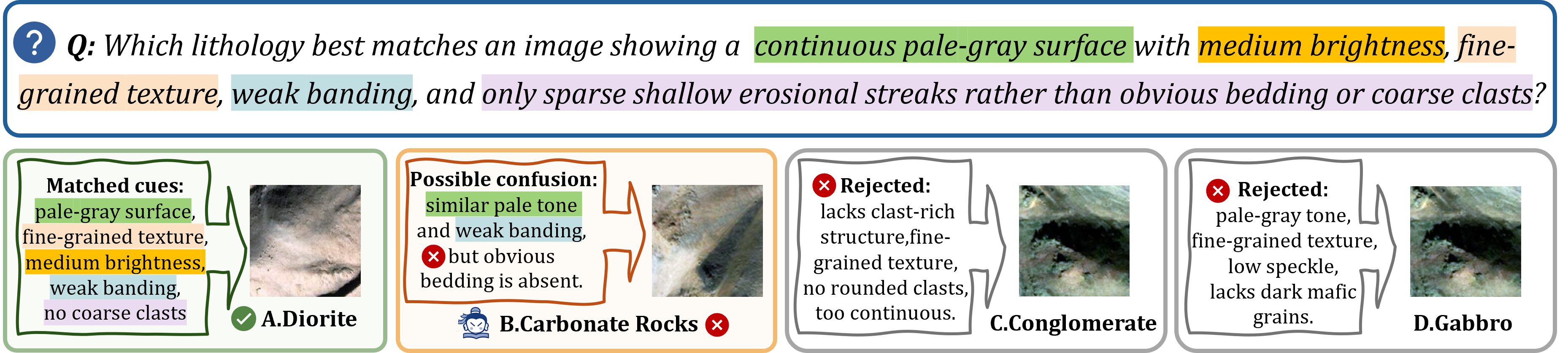}
  \caption{Visual reasoning example for a LithoBench multiple-choice question.}
\vspace{-14pt}
\label{fig:case_study}
\end{figure}
As shown in Fig.~\ref{fig:case_study}, we select a multiple-choice question from the Identification and Description category to analyze the misclassification result of InternVL3.5-8B-HF (FT). The correct answer to this question is A: Diorite, whereas the model predicts B: Carbonate rocks. From the retrieved images corresponding to the four options, it can be observed that A and B are highly similar in overall visual appearance: both exhibit relatively continuous light-gray to gray-white exposed surfaces, moderate brightness, fine textures, fewer spots, and no obvious coarse clastic structures. Such similar tones and fine-grained textures can easily cause the model to misclassify the target image as carbonate rocks.

However, the correct option, Diorite, is more consistent with the target image. The target image shows a relatively homogeneous light-gray continuous surface overall, with only weak linear or weak banded patterns. It lacks the coarse-grained clastic features commonly seen in conglomerate and also lacks the darker, more mafic overall tone of gabbro. Meanwhile, compared with many carbonate rocks, this image does not show clear bedding or sedimentary banding structures. Therefore, this sample is more consistent with the remote sensing visual characteristics of diorite.

This case indicates that the model error is not random, but is influenced by visually similar distractor options. The image RAG mechanism in this paper can retrieve candidate images that are highly similar to the correct answer in tone, texture, and spatial morphology, thereby constructing options with real visual ambiguity. This demonstrates that the multiple-choice questions in \lith not only evaluate the model’s category recognition ability, but also further examine its ability to distinguish fine-grained lithological differences and diagnostic geological features.

Existing remote-sensing vision-language datasets usually obtain relatively high automatic evaluation scores. On the UCM-Captions dataset, RSGPT\cite{hu2025rsgpt}  achieves BLEU-4 and ROUGE-L scores of 0.657 and 0.783, respectively, while GeoChat\cite{kuckreja2024geochat} further improves these two scores to 0.807 and 0.791. On the RSICD dataset, although the scores are lower than those on UCM-Captions, GeoChat still achieves BLEU-4 and ROUGE-L scores of 0.574 and 0.563, respectively. In contrast, even after domain-specific fine-tuning, the best result of current models on the open-ended task of \lith reaches only 0.108 in BLEU-4 and 0.257 in ROUGE-L, which is substantially lower than those on existing remote-sensing vision-language datasets.

This performance gap indicates, on the one hand, that \lith is not a simple general-purpose remote-sensing scene description dataset, but a high-quality and highly challenging benchmark constructed for expert-level lithological interpretation. Its task design involves fine-grained visual discrimination, geological terminology, lithological knowledge, and multi-step reasoning, thereby providing stronger professionalism and discriminative capability. On the other hand, the results also show that although current vision-language models have achieved promising performance on general remote-sensing scene understanding tasks, they still exhibit clear limitations when facing complex lithology recognition, professional geological semantic understanding, and reasoning-oriented question answering. Therefore, they are not yet capable of effectively handling such expert-level remote-sensing geological understanding tasks. Consequently, the relatively low automatic evaluation scores on \lith do not imply poor data quality; rather, they further demonstrate that \lith can effectively reveal the capability boundaries of existing models in professional geoscience scenarios and provide a more rigorous and valuable evaluation platform for the future development of remote-sensing geological large models.

\section{Conclusion and Future Work}

In this work, we introduce \lith, a multi-level benchmark for remote sensing lithology understanding. Unlike conventional remote sensing datasets that mainly evaluate category recognition, \lith covers five capability levels: Identification and Description, Comparative Analysis, Mechanism Explanation, Practical Application, and Comprehensive Reasoning. This design shifts evaluation from \textit{label prediction} to \textit{geoscientific understanding} of remote sensing imagery. We build \lith through a complete pipeline, including image preprocessing, structured description generation, dual-modal knowledge base construction, multi-level instance generation, quality filtering, and benchmark construction. We further evaluate multiple open-source and closed-source large vision-language models. The results show that \lith can distinguish model capabilities in both basic recognition and higher-order reasoning. Current models show initial lithology recognition ability, but remain limited in Mechanism Explanation, Knowledge Transfer, and Multi-step Geoscientific Reasoning. In future work, we will extend domain-specific fine-tuning experiments and develop an agent system for real-world geological remote sensing tasks. This will further improve model usability in practical lithology interpretation scenarios.

\begin{ack}
This work was supported by UK Research and Innovation (UKRI) under the UK government’s Horizon Europe funding guarantee [grant number EP/Z534511/1].
\end{ack}
\bibliographystyle{plain}
\bibliography{citation}

\vspace{-0.35cm}

\medskip

{
\small

\newpage
\section*{NeurIPS Paper Checklist}

\begin{enumerate}

\item {\bf Claims}
    \item[] Question: Do the main claims made in the abstract and introduction accurately reflect the paper's contributions and scope?
    \item[] Answer: \answerYes{} 
    \item[] Justification: We claim our contribution in the abstract first roughly. Then, we explain their details in a specific paragraph at the of introduction.

\item {\bf Limitations}
    \item[] Question: Does the paper discuss the limitations of the work performed by the authors?
    \item[] Answer: \answerYes{} 
    \item[] Justification: We add a section to discuss the limitation of our work in Sec.~\ref{sec:limi}.

\item {\bf Theory assumptions and proofs}
    \item[] Question: For each theoretical result, does the paper provide the full set of assumptions and a complete (and correct) proof?
    \item[] Answer: \answerNA{} 
    \item[] Justification: The work is a benchmark dataset. We evaluate the performance of main stream VLMs and analyze these results.

    \item {\bf Experimental result reproducibility}
    \item[] Question: Does the paper fully disclose all the information needed to reproduce the main experimental results of the paper to the extent that it affects the main claims and/or conclusions of the paper (regardless of whether the code and data are provided or not)?
    \item[] Answer: \answerYes{} 
    \item[] Justification: We present the experimental results in Sec.~\ref{sec:bench_perform} in details.

\item {\bf Open access to data and code}
    \item[] Question: Does the paper provide open access to the data and code, with sufficient instructions to faithfully reproduce the main experimental results, as described in supplemental material?
    \item[] Answer: \answerYes{} 
    \item[] Justification: We provide open access to the proposed benchmark dataset by attaching a anonymous github link in abstract.

\item {\bf Experimental setting/details}
    \item[] Question: Does the paper specify all the training and test details (e.g., data splits, hyperparameters, how they were chosen, type of optimizer) necessary to understand the results?
    \item[] Answer: \answerYes{} 
    \item[] Justification: Experimental setting details about the work is introduced in Sec.~\ref{sec:experiement_setup}.

\item {\bf Experiment statistical significance}
    \item[] Question: Does the paper report error bars suitably and correctly defined or other appropriate information about the statistical significance of the experiments?
    \item[] Answer: \answerYes{} 
    \item[] Justification: We explain the evaluation metrics in Sec.~\ref{sec:experiement_metrics}, and analyze the statistical significance of the experiments in Sec.~\ref{sec:results_analysis}.

\item {\bf Experiments compute resources}
    \item[] Question: For each experiment, does the paper provide sufficient information on the computer resources (type of compute workers, memory, time of execution) needed to reproduce the experiments?
    \item[] Answer: \answerYes{} 
    \item[] Justification: We add computer resource information in Sec.~\ref{sec:experiement_setup}.
    
\item {\bf Code of ethics}
    \item[] Question: Does the research conducted in the paper conform, in every respect, with the NeurIPS Code of Ethics \url{https://neurips.cc/public/EthicsGuidelines}?
    \item[] Answer: \answerYes{} 
    \item[] Justification: We verify that each aspect of the work is conducted with the NeurIPS Code of Ethics.

\item {\bf Broader impacts}
    \item[] Question: Does the paper discuss both potential positive societal impacts and negative societal impacts of the work performed?
    \item[] Answer: \answerYes{} 
    \item[] Justification: We discuss the border impact of the work in Sec.~\ref{sec:B_impact}.
    
\item {\bf Safeguards}
    \item[] Question: Does the paper describe safeguards that have been put in place for responsible release of data or models that have a high risk for misuse (e.g., pre-trained language models, image generators, or scraped datasets)?
    \item[] Answer: \answerNA{} 
    \item[] Justification: The work adopts open-source earth observation data for scientific research without risk for misuse.

\item {\bf Licenses for existing assets}
    \item[] Question: Are the creators or original owners of assets (e.g., code, data, models), used in the paper, properly credited and are the license and terms of use explicitly mentioned and properly respected?
    \item[] Answer: \answerYes{} 
    \item[] Justification: We have checked licensed of adapted models and data, and confirm that our work obeys the above rules.

\item {\bf New assets}
    \item[] Question: Are new assets introduced in the paper well documented and is the documentation provided alongside the assets?
    \item[] Answer: \answerNA{} 
    \item[] Justification: \justificationTODO{}

\item {\bf Crowdsourcing and research with human subjects}
    \item[] Question: For crowdsourcing experiments and research with human subjects, does the paper include the full text of instructions given to participants and screenshots, if applicable, as well as details about compensation (if any)? 
    \item[] Answer: \answerNA{} 
    \item[] Justification: \justificationTODO{}

\item {\bf Institutional review board (IRB) approvals or equivalent for research with human subjects}
    \item[] Question: Does the paper describe potential risks incurred by study participants, whether such risks were disclosed to the subjects, and whether Institutional Review Board (IRB) approvals (or an equivalent approval/review based on the requirements of your country or institution) were obtained?
    \item[] Answer: \answerNA{} 
    \item[] Justification: \justificationTODO{}

\item {\bf Declaration of LLM usage}
    \item[] Question: Does the paper describe the usage of LLMs if it is an important, original, or non-standard component of the core methods in this research? Note that if the LLM is used only for writing, editing, or formatting purposes and does \emph{not} impact the core methodology, scientific rigor, or originality of the research, declaration is not required.
    \item[] Answer: \answerYes{} 
    \item[] Justification: We claim the LLM usage in Sec.~\ref{sec:LLM_Usage}.

\end{enumerate}

\clearpage
\appendix
\etocdepthtag.toc{mtappendix}

\vspace*{1mm}
\begin{center}
    \LARGE \bfseries Appendix of \lith
\end{center}

\etocsettagdepth{mtchapter}{none}
\etocsettagdepth{mtappendix}{subsection}

{
  \hypersetup{
    colorlinks=true,
    linkcolor=kleinblue
  }

  \begingroup
  \makeatletter

  \let\oldl@section\l@section
  \renewcommand{\l@section}[2]{%
    \vspace{-5pt}%
    \oldl@section{#1}{#2}%
    \vspace{5pt}%
  }

  \let\oldl@subsection\l@subsection
  \renewcommand{\l@subsection}[2]{%
    \oldl@subsection{#1}{#2}%
    \vspace{5pt}%
  }

  \makeatother

  \tableofcontents
  \endgroup
}

\newpage

\section{Related Works}
\label{sec:related_works}

Remote sensing lithology datasets are fundamental to lithology recognition, geological mapping, and resource exploration~\cite{peyghambari2021hyperspectral,hajaj2024review,shebl2023prisma,ghoneim2024machine}. 
Existing lithology-related studies mainly focus on classification or geological unit mapping. 
For example, WorldView-3 imagery has been used for lithological mapping and compared with ASTER and Landsat-8 OLI data~\cite{ye2017assessment}; TASI thermal infrared hyperspectral data have been used for CNN-based lithology classification~\cite{liu2021lithology}; and ZY1-02D hyperspectral imagery has been combined with XGBoost for lithology classification~\cite{lin2023lithological}. 
PRISMA-based studies also demonstrate the value of hyperspectral data and machine learning for lithological mapping~\cite{shebl2023prisma}. 
However, these datasets are usually supervised by category labels or mapping results, lacking question-answer annotations, open-ended geological explanations, and multi-level reasoning tasks.

Beyond lithology-specific datasets, many remote sensing benchmarks have been developed for land-cover classification, object detection, segmentation, and geospatial foundation model evaluation, as summarized in Table~\ref{tab:benchmark_comparison}. 
Representative examples include HyRANK for hyperspectral land-cover classification~\cite{karantzalos2018hyrank,christovam2019land}, OpenEarthMap for global land-cover segmentation~\cite{xia2023openearthmap}, FAIR1M for fine-grained object detection~\cite{sun2022fair1m}, and SAMRS for large-scale segmentation and detection~\cite{wang2023samrs}. 
Geo-Bench and PANGAEA further extend evaluation to multiple geospatial tasks and sensors~\cite{lacoste2023geo,marsocci2024pangaea}. 
Nevertheless, these benchmarks mainly evaluate visual recognition or pixel/object-level perception, and rarely examine lithological evidence interpretation, geological comparison, or mechanism-oriented reasoning.

Recent remote sensing vision-language benchmarks introduce captioning, grounding, and visual question answering. 
RSGPT/RSICap focuses on remote sensing captioning~\cite{hu2025rsgpt}; EarthVQA, VRSBench, and RSVLM-QA provide remote sensing VQA tasks with different question types~\cite{wang2024earthvqa,li2024vrsbench,zi2025rsvlm}; and broader benchmarks such as OmniEarth and MSEarth evaluate geospatial or Earth science multimodal understanding~\cite{Fu2026OmniEarth,zhao2025msearth}. 
Despite these advances, most existing vision-language benchmarks target generic remote sensing or broad Earth science scenarios rather than expert-level lithology interpretation.

In contrast, \lith is specifically designed for remote sensing lithology interpretation. 
It contains 112,163 lithology-related image patches and 10,000 final question-answer pairs across 12 representative lithology and land-cover categories. 
Different from existing classification, mapping, captioning, or general VQA benchmarks, \lith evaluates models through both multiple-choice and open-ended tasks organized into five geological capability levels, enabling a systematic assessment of lithology identification, visual evidence description, comparative analysis, mechanism explanation, practical application, and comprehensive reasoning.

\begin{table*}[t]
\centering
\caption{Comparison with representative remote-sensing and geoscience benchmarks.}
\label{tab:benchmark_comparison}
\scriptsize
\setlength{\tabcolsep}{3pt}
\resizebox{\textwidth}{!}{
\begin{tabular}{l c c c c c l}
\toprule
\textbf{Dataset} 
& \textbf{Field} 
& \textbf{Dataset Type} 
& \textbf{Visual Samples} 
& \textbf{Supervision} 
& \textbf{Types} 
& \textbf{Tasks} \\
\midrule

HyRANK~\cite{karantzalos2018hyrank} 
& RS 
& HSI classification 
& 5 scenes 
& Pixel labels 
& 14 classes 
& Land-cover classification \\

WorldView-3 Lithology~\cite{ye2017assessment} 
& Litho-RS 
& Lithology mapping 
& RS scenes 
& Lithology labels 
& 1 task 
& Lithological mapping \\

TASI Lithology~\cite{liu2021lithology} 
& Litho-RS 
& Lithology classification 
& 3 sites 
& Lithology labels 
& 1 task 
& Lithology classification \\

PRISMA Lithology~\cite{shebl2023prisma} 
& Litho-RS 
& Lithology mapping 
& RS scenes 
& Lithology labels 
& 1 task 
& Lithological mapping \\

OpenEarthMap~\cite{xia2023openearthmap} 
& RS 
& Segmentation 
& 5,000 images 
& 2.2M labels 
& 8 classes 
& Land-cover segmentation \\

FAIR1M~\cite{sun2022fair1m} 
& RS 
& Detection 
& $>$15,000 images 
& $>$1M labels 
& 37 classes 
& Fine-grained object detection \\

SAMRS~\cite{wang2023samrs} 
& RS 
& Seg./Det. 
& 105,090 images 
& 1.67M labels 
& 3 tasks 
& Segmentation and detection \\

Geo-Bench~\cite{lacoste2023geo} 
& Geo-RS 
& GFM evaluation 
& Multiple datasets 
& Task labels 
& 12 tasks 
& Classification and segmentation \\

PANGAEA~\cite{marsocci2024pangaea} 
& Geo-RS 
& GFM evaluation 
& Multiple datasets 
& Task labels 
& Multiple tasks 
& Multi-sensor geospatial evaluation \\

RSGPT / RSICap~\cite{hu2025rsgpt} 
& RS-VL 
& Captioning 
& RS images 
& 2,585 captions 
& 1 task 
& Remote-sensing captioning \\

EarthVQA~\cite{wang2024earthvqa} 
& RS-VQA 
& VQA 
& 6,000 images 
& 208,593 QA 
& 4 Q-types 
& Judging, counting, situation, analysis \\

VRSBench~\cite{li2024vrsbench} 
& RS-VL 
& VL benchmark 
& 29,614 images 
& 123,221 QA 
& 3 tasks 
& Captioning, grounding, VQA \\

RSVLM-QA~\cite{zi2025rsvlm} 
& RS-VQA 
& VQA 
& 13,820 images 
& 162,373 QA 
& 6 Q-types 
& Caption-based and counting QA \\

OmniEarth~\cite{Fu2026OmniEarth} 
& Geo-VL 
& VLM evaluation 
& 9,275 images 
& 44,210 instructions 
& 28 tasks 
& Perception, reasoning, robustness \\

MSEarth~\cite{zhao2025msearth} 
& Earth Sci. 
& Scientific VLM 
& $>$289K figures 
& QA / captions 
& 3 tasks 
& Captioning, MCQ, open reasoning \\

\textbf{\lith(Ours)}
& \textbf{Litho-RS}
& \textbf{Lithology VQA}
& \textbf{112,163 patches}
& \textbf{10,000 QA}
& \textbf{5 levels}
& \textbf{ID,Cmp,Exp,App,Rea} \\

\bottomrule
\end{tabular}
}
\vspace{1mm}
\begin{flushleft}
\scriptsize
\textit{Notes.} RS: remote sensing; Litho-RS: remote-sensing lithology; Geo-RS: geospatial remote sensing; 
RS-VL: remote-sensing vision-language; RS-VQA: remote-sensing visual question answering; 
HSI: hyperspectral imagery; GFM: geospatial foundation model; QA: question-answer pairs.
\end{flushleft}
\end{table*}

\section{Dataset Construction Details}

\subsection{Image Source and Patch Preprocessing}
\label{sec:image_source_preprocessing}
The raw imagery used for constructing \lith was collected from four Gaofen-2 (GF-2) satellite scenes over exposed surface regions in northwestern China. The study area contains complex geological structures and diverse lithology-related surface materials. We define twelve representative categories, including arenaceous rocks, breccia, carbonate rocks, conglomerate, crushed stone, diorite, gabbro, granite, gravel, slate, soil, and water. Representative examples are shown in Fig.~\ref{fig:pipeline}(B). To improve label reliability, the category labels were not assigned from a single source. Instead, they were cross-calibrated using field geological survey records, core sampling information, and remote-sensing interpretation results, so as to reduce label ambiguity caused by visually similar but geologically different surface materials.The lithology masks were produced by integrating field geological survey records, core sampling information, and expert remote-sensing interpretation, and were then aligned with the GF-2 scenes for patch-level cropping.

The original GF-2 scenes and their corresponding lithology masks were read as raster arrays using GDAL. Since the original images have large spatial extents and high spatial resolution, we cropped each image and its mask into paired $224 \times 224$ patches using a non-overlapping sliding-window strategy. For image storage, each spectral band was linearly normalized to the range of $[0,255]$, and the first three bands were saved as RGB PNG images; the corresponding mask patches were saved as single-band grayscale PNG files. Patches whose mask values were entirely equal to the NoData label were discarded, where the NoData label was set to $15$ in our implementation. Since a patch may contain mixed surface materials, we further assigned a patch-level label according to the dominant lithology in the mask. Specifically, a patch was retained as a single-lithology sample only when one lithology category accounted for more than 80\% of the valid mask pixels. Patches that did not satisfy this threshold were excluded to reduce label noise caused by mixed lithological boundaries. After removing NoData-only windows, mixed-label patches, low-quality samples, and boundary-noise patches, we obtained 112,163 high-quality image patches, which served as the basic image pool for structured description generation, dual-modal knowledge retrieval, and subsequent question-answer construction.

\subsection{Lithology Categories}
\label{sec:lithology_categories}
The lithology categories in \lith were defined based on field geological investigation, core sampling records, and remote-sensing interpretation results. Instead of following a very fine-grained petrological taxonomy, we used an image-oriented category system that is suitable for GF-2 remote-sensing interpretation. The reason is that many fine-grained geological subtypes cannot be reliably distinguished from optical remote-sensing patches alone. For example, loose angular gravel, rounded gravel, and other gravel-like surface materials often show very similar tones, textures, particle-size patterns, and spatial distributions in GF-2 imagery. Therefore, they were merged into the broader \emph{gravel} category. Similarly, coarse sand, medium sand, and fine sand usually present highly similar surface textures and spectral appearances at the patch scale, so they were grouped into the \emph{arenaceous rocks} category. This merging strategy reduces unnecessary label ambiguity and avoids assigning visually indistinguishable subtypes as separate classes.

Following this principle, we finally organized the dataset into twelve representative categories: arenaceous rocks, breccia, carbonate rocks, conglomerate, crushed stone, diorite, gabbro, granite, gravel, slate, soil, and water. These categories cover both major lithological units and auxiliary surface-cover types that commonly appear in exposed geological regions. Rock categories such as carbonate rocks, slate, granite, diorite, and gabbro were retained because they have relatively stable geological meanings and can often be associated with distinguishable visual or geomorphic patterns. Categories such as soil and water were included because they frequently occur in the same scenes and may affect lithology interpretation. Overall, the category design aims to balance geological validity and remote-sensing separability, ensuring that each class is meaningful for geological interpretation while remaining visually identifiable at the image-patch level.

\subsection{Expert Panel Configuration}
\label{sec:expert_panel}

To ensure the reliability and geological validity of \lith, we organize an expert-supervised annotation and verification team consisting of three senior domain experts, five Ph.D. researchers, and ten graduate students. The expert team is involved throughout the construction process, including data preprocessing, prompt design, model selection, QA generation, and final benchmark filtering. During preprocessing, experts examine GF-2 patches together with lithology masks and geological survey information to remove samples with mixed lithology, unclear boundaries, severe shadows, weak exposure, or inconsistent label-image correspondence. During description and QA generation, experts refine all prompt templates and inspect the generated results to ensure that they follow the expected format, cover lithology-related cues, and remain geologically meaningful. For model selection, we jointly consider LLM-based evaluation results and real expert assessment to choose models that produce more reliable descriptions and QA instances. In the final filtering stage, experts review the 15,000 high-scoring candidates from multiple perspectives, including geological correctness, image-question alignment, answerability, reasoning validity, terminology appropriateness, distractor plausibility, and leakage risk. After this expert-guided verification, 10,000 instances are selected as the official benchmark.

\subsection{Structured Description Generation}

After obtaining the GF-2 image patches, we generated structured geological descriptions as intermediate representations for subsequent knowledge retrieval and question-answer construction. Before large-scale generation, we first compared the description quality of different LVLMs on the same lithological images. Given the original user instruction, the input image, and two candidate descriptions generated by Assistant A and Assistant B, a judge model was asked to compare the two responses and rank them from best to worst. The evaluation considered instruction adherence, visual relevance, geological accuracy, completeness, description granularity, and objectivity. As shown in Fig.~\ref{fig:prompt_pairwise_judgement}, the judge prompt required a valid JSON output containing both an analysis field and a final judgement field, so that the pairwise comparison results could be automatically parsed and summarized. Based on this pairwise evaluation, GPT-5.4 showed better overall description quality than the compared alternatives, and was therefore selected as the default model for subsequent structured description generation and question-answer construction.

For each retained image patch, we then used a structured prompt to generate a geological description. Instead of producing free-form captions, the prompt required the model to output a valid JSON object with predefined fields, including tone and color, texture, pattern and structure, hydro-geomorphological features, and spatial-scale cues. For each field, the model provided both a categorical value and a short evidence phrase grounded in visible image features. This format makes the descriptions more consistent, interpretable, and easier to parse automatically. As shown in Fig.~\ref{fig:prompt_image_description}, the prompt explicitly constrains the content scope and output format, which reduces uncontrolled generation and supports large-scale description generation with a unified structure.

We further conducted quality control for the generated image-description pairs. The quality-control prompt evaluated each sample in two stages. The first stage assessed the visual quality of the image itself, including clarity, texture visibility, contrast and exposure, information content, structural or boundary visibility, and noise or artifact control. The second stage evaluated the generated description with respect to the image evidence, focusing on image-description consistency, coverage of visible features, specificity, evidence grounding, internal consistency, and hallucination control. This design separates image usability from description reliability, preventing low-quality images and inaccurate descriptions from being conflated. As shown in Fig.~\ref{fig:prompt_joint_image_description_qc}, the prompt required a structured JSON output containing image quality, description quality, and an overall recommendation. We ranked all image-description pairs according to the final quality score and retained the top 50\% as high-quality image-description pairs for subsequent knowledge retrieval, question generation, and benchmark construction.

\begin{promptbox}{Prompt for Pairwise Model Response Judgement}
Please serve as an unbiased judge in assessing the quality of the responses from AI assistants regarding the user's instruction and a figure.

Please examine the provided image attentively. Begin by conducting a detailed analysis of the responses provided. Capture your comprehensive observations and insights in the "Analysis" section. Following your analysis, move on to the judgement phase, where you will make informed decisions or conclusions based on the analysis conducted. Give your final judgements in the "Judgement" section. Ensure that your final output in a JSON format with keys "Analysis" for the initial response analysis, and "Judgement" for your final judgement only. Ensure that the content under each key does not contain any nested JSON structures.

[The Start of User Instruction]
Please describe the following features of the image in approximately 500 words:
Tone/Color, Texture, Pattern/Structure, Hydro-Geomorph, Spatial Scale Cues.
Output format: "Field + Enumerated Value + Evidence Phrase".
[The End of User Instruction]

[The Start of Assistant A's Answer]
{response_A}
[The End of Assistant A's Answer]

[The Start of Assistant B's Answer]
{response_B}
[The End of Assistant B's Answer]

You will be presented with responses from two different assistants to the same user instruction.
Your task is to assess and compare these responses based on how effectively they adhere to the user's original instruction and how aptly they address the user's inquiry.
After your assessment and comparison, you should RANK the responses from best to worst as the following template. If Assistant A is the better response and Assistant B is the worse response, you should output like [[A]], [[B]]. Indicate your final rank in the key "Judgement".

Your assessment should identify whether the assistant effectively adheres to the user's instruction and addresses the user's inquiry.
In your evaluation, weigh factors such as relevance, accuracy, comprehensiveness, creativity, and the granularity of the responses.
Do not allow the length of the responses to influence your evaluation.
Do not favor certain names or position of the assistants. Be as objective as possible.
\end{promptbox}

\captionof{figure}{Prompt for pairwise judgement between two model-generated lithological image descriptions.}
\label{fig:prompt_pairwise_judgement}
\begin{promptbox}{Prompt for Structured Image Description}
This is a lithological image captured by the Gaofen-2 remote sensing satellite, showing {label}. Analyze the image and output the visual features STRICTLY as a valid JSON object.

CRITICAL RULES:
1. DO NOT output any internal thinking processes, planning, or reasoning. You are STRICTLY FORBIDDEN from using <think> tags.
2. DO NOT output any conversational text, greetings, or explanations before or after the JSON.
3. Output ONLY a valid, parsable JSON object matching the exact structure below.

Required JSON Structure:
{
  "Tone_Color": {
    "Brightness": {"value": "Dark/Medium/Bright", "evidence": "..."},
    "Hue_Bias": {"value": "Gray/Yellow-brown/Red-brown/Blue-gray", "evidence": "..."},
    "Contrast": {"value": "...", "evidence": "..."}
  },
  "Texture": {
    "Granularity": {"value": "Fine/Medium/Coarse", "evidence": "..."},
    "Homogeneity": {"value": "...", "evidence": "..."},
    "Speckle": {"value": "...", "evidence": "..."},
    "Banding": {"value": "...", "evidence": "..."},
    "Noise_Level": {"value": "...", "evidence": "..."}
  },
  "Pattern_Structure": {
    "Linear_Structure": {"value": "Strong/Weak", "evidence": "..."},
    "Reticulate_Dendritic_Pinnate": {"value": "...", "evidence": "..."},
    "Bedding_Banding": {"value": "...", "evidence": "..."},
    "Circular_Arcuate_shapes": {"value": "...", "evidence": "..."},
    "Blocky_Fragmentation": {"value": "...", "evidence": "..."}
  },
  "Hydro_Geomorph": {
    "Gully_Density": {"value": "High/Low", "evidence": "..."},
    "Slope_Erosion_Patterns": {"value": "...", "evidence": "..."},
    "Fan_shaped_Accumulation": {"value": "...", "evidence": "..."},
    "Signs_of_Collapse_Slumping": {"value": "morphological description only", "evidence": "..."}
  },
  "Spatial_Scale_Cues": {
    "Presence": {"value": "Large Continuous Surfaces/Fragmented Patches", "evidence": "..."},
    "Boundary_Clarity": {"value": "...", "evidence": "..."}
  }
}

Important: If you cannot see or access the image, directly output exactly: {"error": "no picture"}
\end{promptbox}

\captionof{figure}{Prompt for generating structured visual descriptions from GF-2 lithological image patches.}
\label{fig:prompt_image_description}
\begin{promptbox}{Prompt for Joint Image--Description Quality Control}
You are a strict quality-control reviewer for a remote sensing lithology dataset.

Image label: {label}
Water-related label: {is_water_related}

Below is the existing generated description for this image:
<description>
{description_text}
</description>

Your job has TWO stages, in this exact order:

Stage 1 -- Evaluate IMAGE QUALITY only.
Score the image from 0 to 10 for visual usability in lithology interpretation.
Focus on:
- clarity/sharpness
- texture visibility
- contrast/exposure balance
- information content
- structure/boundary visibility
- noise/artifacts

Scoring guidance:
- 0-1: almost unusable, nearly empty, broken, totally black/white, or no useful detail
- 2-3: very poor, severe blur or severe under/overexposure
- 4-5: marginal, limited but some usable information
- 6-7: usable
- 8-9: high quality
- 10: excellent

Important water rule:
If the label is water-related, dark or near-black appearance alone is NOT enough to assign a very low image-quality score.
Judge whether the image still preserves meaningful boundary, tonal variation, spatial pattern, or contextual interpretability.

Stage 2 -- Evaluate DESCRIPTION QUALITY after considering the image.
Score the description from 0 to 10.
Focus on:
- consistency_with_image
- coverage_of_visible_features
- specificity_and_precision
- evidence_grounding
- internal_consistency
- hallucination_control

Description scoring rules:
- Penalize claims that are too specific, too certain, or not visually supported.
- If the image quality is low, a cautious/uncertainty-aware description can still be decent.
- If the image quality is low but the description confidently invents detailed structures, score the description low.
- Description quality is NOT the same as image quality.

Output ONLY one valid JSON object. No markdown, no explanation.

Required JSON schema:
{
  "image_quality": {
    "score": 0,
    "level": "very_poor/poor/fair/good/excellent",
    "usable_for_interpretation": false,
    "dimensions": {
      "clarity_sharpness": 0,
      "texture_visibility": 0,
      "contrast_exposure": 0,
      "information_content": 0,
      "structure_boundary_visibility": 0,
      "noise_artifact_control": 0
    },
    "flags": {
      "almost_black": false,
      "almost_white": false,
      "blurred": false,
      "underexposed": false,
      "overexposed": false,
      "low_texture": false,
      "heavy_noise_or_artifacts": false
    },
    "summary": ""
  },
  "description_quality": {
    "score": 0,
    "level": "very_poor/poor/fair/good/excellent",
    "dimensions": {
      "consistency_with_image": 0,
      "coverage_of_visible_features": 0,
      "specificity_and_precision": 0,
      "evidence_grounding": 0,
      "internal_consistency": 0,
      "hallucination_control": 0
    },
    "major_issues": ["", ""],
    "summary": ""
  },
  "overall_assessment": {
    "final_recommendation": "keep/review/drop",
    "keep_for_training": false,
    "reason": ""
  }
}

If you cannot see or access the image, output exactly:
{"error": "no picture"}
\end{promptbox}

\captionof{figure}{Prompt for joint quality control of remote-sensing image-description pairs.}
\label{fig:prompt_joint_image_description_qc}

\subsection{Dual-modal Retrieval Database Construction}
\label{sec:dual_modal_retrieval_database}

We first build a text RAG database and an image retrieval database to support candidate QA generation. 
For the text RAG database, we collect geoscientific materials from two main sources. 
The first source consists of recent papers published within the past five years and retrieved through Google Scholar searches using keywords related to petrology and remote-sensing geology. 
The second source includes classical domestic and international textbooks in remote sensing, geology, petrology, and geological interpretation. 
These materials provide professional terminology, lithological interpretation principles, and domain-specific geological knowledge. 
All documents are segmented into semantic chunks, encoded into dense vectors, and stored in a vector database. 
During QA generation, the retrieval query is formed by combining the target lithology label, the structured image description, and the target capability level. 
The retrieved knowledge provides causal geological context and interpretation principles, especially for explanation, application, and reasoning questions.

For the image retrieval database, representative high-quality image patches are manually selected by an expert team from the full collection of more than 110,000 image patches. 
These selected samples are required to clearly represent the visual characteristics of their corresponding lithology-related categories and are then encoded into visual embeddings. 
Given a target image, the system retrieves visually similar samples from different categories. 
These retrieved images are not directly exposed in the final benchmark question, but are used internally to construct plausible distractors for multiple-choice questions and to strengthen comparative interpretation. 
This dual-modal retrieval design ensures that the generated questions are grounded in both visual evidence and geological knowledge.

\subsection{Multi-level QA Pair Generation}
\label{sec:multi_level_qa_generation}

\begin{table}[H]
\centering
\caption{Capability-specific guidance used in the multi-level QA generation prompts.}
\label{tab:qa_prompt_guidance}
\small
\begin{tabular}{p{3.7cm}p{9cm}}
\toprule
\textbf{Capability Level} & \textbf{Generation Focus} \\
\midrule
Identification and Description
& Focuses on rock-type recognition, feature identification, and descriptive interpretation of tone/color, texture, structure, hydro-geomorphic expression, and spatial-scale cues. \\

Comparative Analysis
& Focuses on comparative interpretation and discrimination between visually similar lithologies, emphasizing differences in tone/color, texture, structural pattern, spatial continuity, and geomorphic expression. \\

Mechanism Explanation
& Focuses on causal and interpretive questions that connect visual evidence with geological processes, lithological characteristics, and professional geological terminology. \\

Practical Application
& Focuses on practical, assessment-oriented, and decision-oriented questions, such as geological interpretation, engineering relevance, resource implication, or hazard-related assessment. \\

Comprehensive Reasoning
& Focuses on logical deduction, temporal sequencing, geological history inference, and step-by-step reasoning from visual evidence such as deformation, cross-cutting relationships, differential weathering, and depositional indicators. \\
\bottomrule
\end{tabular}
\end{table}
Based on the structured description and the retrieved resources, we generate candidate QA pairs across five capability levels: Identification and Description, Comparative Analysis, Mechanism Explanation, Practical Application, and Comprehensive Reasoning.
For each target image, the system generates five capability-specific candidate questions, with one candidate corresponding to each capability level. 
To balance question format and capability coverage, two of the five capability levels are assigned as multiple-choice questions and the remaining three are assigned as open-ended questions. 
Therefore, each image contributes five candidate QA pairs with an MCQ-to-OPEN ratio of \(2:3\), which is consistent with the final benchmark design of 4,000 MCQ instances and 6,000 OPEN instances. 
The assignment is deterministic rather than purely random: a fixed global seed and the normalized image path are used to initialize a deterministic random generator, ensuring that the same image receives the same question-type assignment across different runs. 
The generated assignment is saved as a JSONL plan file, where each record contains the image path, label, a \texttt{mode\_map} specifying whether each capability level is assigned to MCQ or OPEN, and a \texttt{seed\_map} specifying which question prototype is used for each level. 
The five QA generation workers then read this shared plan file and generate questions according to their assigned capability levels.

Instead of using a single generic question-generation prompt, we design capability-specific guidance and question prototypes for each level. 
This design allows the generated questions to reflect different forms of lithological understanding, ranging from visual recognition and descriptive interpretation to comparative discrimination, causal explanation, practical assessment, and multi-step geological reasoning. 
For multiple-choice questions, visually similar reference images are internally used to construct plausible distractors. 
However, the final question is required to appear as a standalone single-image benchmark item, and no wording is allowed to reveal the existence, position, or layout of the retrieved reference images. 
For open-ended questions, the prompt encourages evidence-grounded answers based on the target image description and, for higher-order levels, retrieved professional knowledge. 
All prompts enforce a strict JSON output format, which facilitates automatic parsing, quality checking, and downstream benchmark construction.

To ensure that the generated QA pairs cover different levels of lithological understanding, we design capability-specific guidance for each question category. 
As summarized in Table~\ref{tab:qa_prompt_guidance}, the five capability levels correspond to different generation focuses, ranging from basic visual identification to comparison, causal explanation, Practical Application, and multi-step geological reasoning. 
This design prevents the generated questions from collapsing into simple category recognition and encourages the benchmark to evaluate both visual perception and higher-order geoscientific reasoning.

The QA generation prompts are implemented as template-based instructions with several input variables. 
These variables specify the target image label, structured image description, retrieved visual references, retrieved text knowledge, capability level, and generation constraints. 
Table~\ref{tab:qa_prompt_variables} lists the main variables used in the prompt templates. 
By explicitly defining these variables, the generation process becomes easier to reproduce, debug, and adapt to different lithology categories or capability levels.

\begin{table}[H]
\centering
\caption{Variables used in the multi-level QA generation prompts.}
\label{tab:qa_prompt_variables}
\small
\begin{tabular}{p{3.8cm}p{8.5cm}}
\toprule
\textbf{Variable} & \textbf{Meaning} \\
\midrule
\texttt{\{target\_label\}} & Lithology or land-cover label of the target image. \\
\texttt{\{target\_description\}} & Structured geological description generated for the target image. \\
\texttt{\{top1\_label\}}, \texttt{\{top2\_label\}}, \texttt{\{top3\_label\}} & Labels of visually similar reference images used internally for distractor design. \\
\texttt{\{retrieved\_knowledge\}} & Top-ranked professional knowledge fragments retrieved from the text knowledge base. \\
\texttt{\{capability\_level\}} & One of the five capability levels: Identification and Description, Comparative Analysis, Mechanism Explanation, Practical Application, and Comprehensive Reasoning. \\
\texttt{\{category\_guidance\}} & Capability-specific generation guidance. \\
\texttt{\{question\_prototypes\}} & Candidate question prototypes used as generation references. \\
\texttt{\{mcq\_focus\}} & Capability-specific focus of the MCQ, such as recognition, comparison, explanation, application, or reasoning. \\
\texttt{\{answer\_length\}} & Recommended answer length for open-ended questions. \\
\bottomrule
\end{tabular}
\end{table}

Based on these capability definitions and input variables, we use two prompt templates to generate multiple-choice and open-ended QA pairs, respectively. 
The MCQ prompt emphasizes plausible distractor construction using visually similar reference images, while the open-ended prompt focuses on evidence-grounded geological interpretation with support from structured descriptions and retrieved professional knowledge. 
As shown in Figs.~\ref{fig:prompt_mcq_generation} and~\ref{fig:prompt_open_generation}, these templates make the QA generation process controllable, reproducible, and easy to parse automatically.

\begin{promptbox}{Prompt for Multi-level MCQ Generation}
You are a geologic remote-sensing expert creating English {capability_level}-oriented QA pairs for a high-quality lithology image dataset.

You will see a 2$\times$2 composite image:
- Top-left: target image (the question must be about this image)
- Top-right: top1 similar reference image
- Bottom-left: top2 similar reference image
- Bottom-right: top3 similar reference image

[TARGET IMAGE LABEL]
{target_label}

[REFERENCE IMAGE LABELS]
top1 = {top1_label}
top2 = {top2_label}
top3 = {top3_label}

[TARGET IMAGE DESCRIPTION]
{target_description}

[RETRIEVED PROFESSIONAL KNOWLEDGE]
{retrieved_knowledge}

[CURRENT QUESTION CATEGORY]
{capability_level}

[CATEGORY-SPECIFIC GUIDANCE]
{category_guidance}

[QUESTION PROTOTYPES]
{question_prototypes}

[GENERAL RULES]
1. The generated question must be about the target image only.
2. Any internally provided reference images may be used only as hidden support for distractor design, comparative reasoning, or quality control.
3. The final question, answer, and explanation must be written exactly as a single-image benchmark item.
4. The final wording must read naturally as if the user is seeing only one standalone target image.
5. Never mention image layout, image positions, panel structure, or multi-image composition in any form.
6. Never use expressions such as "top-left", "top-right", "bottom-left", "bottom-right", "left image", "right image", "reference image", "similar image", "another image", "another panel", "compared with the reference image", or any equivalent wording.
7. Do not say or imply that multiple images were provided.
8. Do not explicitly mention any hidden support images even when using them internally.
9. You may draw inspiration from any of the provided question prototypes, or combine ideas from multiple prototypes.
10. Do not copy any prototype verbatim.
11. Keep the question professional and geologically meaningful.
12. Keep the answer consistent with the target image evidence and any internally used support information, while referring only to the target image in the final wording.
13. Output strictly valid JSON only. No markdown. No extra text.

Generate exactly 1 English single-choice multiple-choice question.

Requirements:
1. Provide exactly 4 options labeled A/B/C/D.
2. There must be exactly 1 correct answer.
3. The question should be {mcq_focus} rather than a purely superficial lookup.
4. Distractors must be plausible, domain-relevant, and challenging.
5. Use retrieved professional knowledge when available to improve correctness and professionalism, but do not copy it mechanically.
6. Do not leak the correct answer in the question stem.

Output JSON:
{
  "question": "...",
  "options": [
    {"key": "A", "text": "..."},
    {"key": "B", "text": "..."},
    {"key": "C", "text": "..."},
    {"key": "D", "text": "..."}
  ],
  "answer_key": "A",
  "answer_text": "...",
  "explanation": "..."
}
\end{promptbox}

\captionof{figure}{Prompt template for generating multiple-choice questions across the five capability levels.}
\label{fig:prompt_mcq_generation}

\begin{promptbox}{Prompt for Multi-level Open-ended QA Generation}
You are a geologic remote-sensing expert creating English {capability_level}-oriented QA pairs for a high-quality lithology image dataset.

You will see one target image only.
The generated question must be about this target image.

[TARGET IMAGE LABEL]
{target_label}

[TARGET IMAGE DESCRIPTION]
{target_description}

[RETRIEVED PROFESSIONAL KNOWLEDGE]
{retrieved_knowledge}

[CURRENT QUESTION CATEGORY]
{capability_level}

[CATEGORY-SPECIFIC GUIDANCE]
{category_guidance}

[QUESTION PROTOTYPES]
{question_prototypes}

[GENERAL RULES]
1. The generated question must focus only on the target image.
2. Do not use any reference images, reference labels, or reference descriptions.
3. Use the retrieved professional knowledge to strengthen the geological interpretation, practical assessment, logical deduction, and professional terminology when available.
4. The answer must remain consistent with the actual image evidence and the target description.
5. If the retrieved knowledge includes general geological principles, adapt them carefully to the current image instead of copying them directly.
6. Do not copy any prototype or retrieved chunk verbatim.
7. Output strictly valid JSON only. No markdown. No extra text.

Generate exactly 1 English open-ended question.

Requirements:
1. The question must be clear, professional, and answerable.
2. The answer must be grounded in the target image evidence, the provided target description, and the retrieved professional knowledge when available.
3. The answer should not rely on unsupported speculation.
4. The answer should reflect expert-level lithological interpretation rather than shallow description only.
5. Recommended answer length: about {answer_length} words.

Output JSON:
{
  "question": "...",
  "answer": "...",
  "explanation": "..."
}
\end{promptbox}

\captionof{figure}{Prompt template for generating open-ended questions across the five capability levels.}
\label{fig:prompt_open_generation}

\subsection{Official Benchmark Selection}
\label{sec:official_benchmark_selection}

After candidate QA generation, we further apply a two-stage selection process to construct the official \lith benchmark. 
The first stage performs LLM-as-a-Judge~\cite{zheng2023judging} quality scoring for all candidate QA pairs. 
Instead of evaluating each question independently from scratch, we group all questions belonging to the same target image and evaluate them together with the shared image and the shared structured description. 
This design reduces repeated visual reasoning and allows the evaluator to compare multiple candidate questions under the same visual evidence. 
For each question, the evaluator assigns scores for question quality, image relevance, clarity, geological soundness, answerability, analysis quality, overall quality, and benchmark leakage risk. 
For multiple-choice questions, distractor quality is additionally scored. 
As shown in Fig.~\ref{fig:prompt_official_selection_scoring}, the scoring prompt explicitly requires discriminative scoring and discourages inflated scores, which is important for downstream ranking and quota-constrained selection.

\begin{promptbox}{Prompt for Candidate QA Quality Scoring}
You are an expert evaluator for a lithology QA benchmark.

You will evaluate a batch of questions that all belong to the SAME target image.
These questions may come from different category files and different modes (mcq/open), but they share the same image and image description.
Please score EACH question independently, while using the shared image evidence and shared image description to reduce repeated reasoning.

Important:
- Use a 10-point scale for all numeric scores.
- You may use one decimal place.
- Do NOT inflate scores.
- Most ordinary questions should fall between 4.0 and 7.5.
- Scores of 8.0-8.4 should indicate clearly above-average items.
- Scores of 8.5 or above should be rare.
- Scores of 9.0 or above should be reserved only for exceptional benchmark items.
- If a question is generic, weakly image-grounded, easy, repetitive, or only moderately useful, it should NOT receive a high score.
- Score separation is essential for downstream ranking, so do not cluster all questions near the top.
- Evaluate the questions comparatively within this batch as well as absolutely.
- Only return scores in JSON format. Do not return explanations, comments, labels, or revision suggestions.

[SHARED IMAGE METADATA]
image_path: {image_path}

[SHARED IMAGE DESCRIPTION]
{image_description}

[QUESTION BATCH]
{questions_block}

Scoring dimensions for EACH question:

1. question_quality_score:
   Evaluate whether the question is well-formed, precise, meaningful, non-trivial, and suitable as a benchmark question.

2. image_relevance_score:
   Evaluate how strongly the question is grounded in and relevant to the target image.

3. clarity_score:
   Evaluate whether the wording is clear, precise, and unambiguous.

4. geological_soundness_score:
   Evaluate whether the geological content is professional, scientifically reasonable, and domain-appropriate.

5. answerability_score:
   Evaluate whether the question can actually be answered from the image and available evidence.

6. distractor_quality_score:
   Only for multiple-choice questions.
   Evaluate whether the incorrect options are plausible, domain-relevant, and sufficiently challenging.
   Use null for open-ended questions.

7. analysis_quality_score:
   Evaluate the quality of the provided answer and analysis.
   Penalize shallow, generic, weakly justified, or weakly image-grounded analysis.

8. overall_score:
   Evaluate the overall quality of the complete QA item by jointly considering the question, the provided answer, the analysis, and the match between the question and the image.
   This score should be discriminative and suitable for ranking.

9. benchmark_leak_risk_score:
   Use a 10-point risk scale.
   1.0 means almost no benchmark leakage risk.
   10.0 means very severe benchmark leakage risk.
   Assign a high score if the item reveals image layout, hidden references, or multi-image setup.

Return STRICTLY VALID JSON ONLY in the following format:
{
  "group_summary": {
    "image_path": "{image_path}",
    "question_count": {question_count}
  },
  "question_evaluations": [
    {
      "qid": 1,
      "question_quality_score": 1.0,
      "image_relevance_score": 1.0,
      "clarity_score": 1.0,
      "geological_soundness_score": 1.0,
      "answerability_score": 1.0,
      "distractor_quality_score": 1.0,
      "analysis_quality_score": 1.0,
      "overall_score": 1.0,
      "benchmark_leak_risk_score": 1.0
    }
  ]
}
\end{promptbox}

\captionof{figure}{Prompt for scoring candidate QA pairs before official benchmark selection.}
\label{fig:prompt_official_selection_scoring}

Based on the above scores, we compute a weighted score for each candidate QA pair. 
Since a higher benchmark leakage risk score indicates a more severe risk, we first convert it into a low-leakage bonus:
\begin{equation}
\mathrm{LLB} = 11 - \mathrm{LeakRisk}.
\end{equation}
For multiple-choice questions, the weighted score is defined as:
\begin{equation}
\begin{aligned}
\mathrm{Score}_{\mathrm{MCQ}} =\;&
0.24\,\mathrm{OQ}
+0.20\,\mathrm{QQ}
+0.17\,\mathrm{IR}
+0.08\,\mathrm{CL}
+0.08\,\mathrm{GS} \\
&+0.07\,\mathrm{AN}
+0.06\,\mathrm{AQ}
+0.07\,\mathrm{DQ}
+0.03\,\mathrm{LLB},
\end{aligned}
\end{equation}
where OQ, QQ, IR, CL, GS, AN, AQ, DQ, and LLB denote overall quality, question quality, image relevance, clarity, geological soundness, answerability, analysis quality, distractor quality, and low-leakage bonus, respectively.

For open-ended questions, the weighted score is defined as:
\begin{equation}
\begin{aligned}
\mathrm{Score}_{\mathrm{OPEN}} =\;&
0.26\,\mathrm{OQ}
+0.22\,\mathrm{QQ}
+0.18\,\mathrm{IR}
+0.08\,\mathrm{CL}
+0.08\,\mathrm{GS} \\
&+0.08\,\mathrm{AN}
+0.07\,\mathrm{AQ}
+0.03\,\mathrm{LLB}.
\end{aligned}
\end{equation}

After scoring, all candidates are divided into ten buckets according to their capability level and question format, namely five capability levels $\times$ two modes. 
Within each bucket, if multiple candidates are generated from the same image, we keep only the candidate with the highest weighted score. 
This local deduplication step avoids over-representing a single image within the same capability-format setting. 
The remaining candidates in each bucket are then sorted by weighted score, with overall quality, question quality, and image relevance used as tie-breaking indicators.

The final benchmark is selected through expert review under strict quota constraints.
For each capability level, we select 800 multiple-choice questions and 1,200 open-ended questions, resulting in 4,000 MCQ instances and 6,000 OPEN instances in total. 
During selection, we also enforce global image uniqueness, so each final QA instance corresponds to a unique image. 
To satisfy all quotas under this global uniqueness constraint, we use a dynamic pressure-based allocation strategy. 
For each unfinished bucket, the pressure is computed as:
\begin{equation}
\mathrm{Pressure}_{b} =
\frac{\mathrm{RemainingQuota}_{b}}
{\mathrm{RemainingAvailableCandidates}_{b}}.
\end{equation}
At each iteration, buckets with higher pressure are prioritized, so that resource-constrained buckets are filled earlier. 
This strategy balances quality ranking, quota satisfaction, category coverage, question-format balance, and image diversity, yielding the official \lith benchmark.

Table~\ref{tab:official_selection_count} summarizes the number of questions in the large-scale candidate pool and the final \lith benchmark across different cognitive levels and question modes. Table~\ref{tab:official_selection_quality} compares the weighted-score statistics before and after official benchmark selection. Overall, the mean weighted score increases from 7.8761 in the original candidate set to 8.2924 in \lith, while the standard deviation decreases from 0.5016 to 0.1281. The same trend is observed for both MCQ and OPEN questions and across all five capability levels. These results show that the final benchmark is not only higher-scoring on average, but also more stable in quality, demonstrating the effectiveness of the LLM-based scoring and quota-constrained dynamic selection strategy.

\begin{table*}[t]
  \caption{Comparison of question counts before and after filtering.}
  \label{tab:official_selection_count}
  \centering
  \resizebox{\textwidth}{!}{
  \begin{tabular}{lcccccccccccccccccc}
    \toprule
    \multirow{2}{*}{Metric}
    & \multicolumn{3}{c}{Overall}
    & \multicolumn{5}{c}{Level-wise}
    & \multicolumn{10}{c}{Level--Mode} \\
    \cmidrule(lr){2-4}
    \cmidrule(lr){5-9}
    \cmidrule(lr){10-19}
    & All & MCQ & Open
    & ID & Cmp & Exp & App & Rea
    & ID-MCQ & ID-Open
    & Cmp-MCQ & Cmp-Open
    & Exp-MCQ & Exp-Open
    & App-MCQ & App-Open
    & Rea-MCQ & Rea-Open \\
    \midrule

    Original Count
    & 240513 & 96104 & 144409
    & 53944 & 52698 & 45062 & 44416 & 44393
    & 21368 & 32576
    & 21203 & 31495
    & 18241 & 26821
    & 17677 & 26739
    & 17615 & 26778 \\

    \lith Count
    & \textbf{10000} & \textbf{4000} & \textbf{6000}
    & \textbf{2000} & \textbf{2000} & \textbf{2000} & \textbf{2000} & \textbf{2000}
    & \textbf{800} & \textbf{1200}
    & \textbf{800} & \textbf{1200}
    & \textbf{800} & \textbf{1200}
    & \textbf{800} & \textbf{1200}
    & \textbf{800} & \textbf{1200} \\

    \bottomrule
  \end{tabular}
  }
\end{table*}

\begin{table*}[t]
  \caption{Comparison of weighted quality scores before and after filtering. Std. denotes the standard deviation of weighted scores, and Mean denotes the average weighted score.}
\label{tab:official_selection_quality}
  \centering
  \resizebox{\textwidth}{!}{
  \begin{tabular}{lcccccccccccccccccc}
    \toprule
    \multirow{2}{*}{Metric}
    & \multicolumn{3}{c}{Overall}
    & \multicolumn{5}{c}{Level-wise}
    & \multicolumn{10}{c}{Level--Mode} \\
    \cmidrule(lr){2-4}
    \cmidrule(lr){5-9}
    \cmidrule(lr){10-19}
    & All & MCQ & Open
    & ID & Cmp & Exp & App & Rea
    & ID-MCQ & ID-Open
    & Cmp-MCQ & Cmp-Open
    & Exp-MCQ & Exp-Open
    & App-MCQ & App-Open
    & Rea-MCQ & Rea-Open \\
    \midrule
    Original Std.
    & 0.5016 & 0.4913 & 0.5080
    & 0.4655 & 0.5377 & 0.4437 & 0.4628 & 0.4526
    & 0.4641 & 0.4105
    & 0.4391 & 0.5458
    & 0.4261 & 0.4543
    & 0.4442 & 0.4721
    & 0.4541 & 0.4516 \\

    \lith Std.
    & \textbf{0.1281} & \textbf{0.1401} & \textbf{0.1162}
    & \textbf{0.1886} & \textbf{0.0804} & \textbf{0.0490} & \textbf{0.0751} & \textbf{0.0502}
    & \textbf{0.0689} & \textbf{0.0790}
    & \textbf{0.0420} & \textbf{0.0646}
    & \textbf{0.0429} & \textbf{0.0513}
    & \textbf{0.0602} & \textbf{0.0707}
    & \textbf{0.0464} & \textbf{0.0481} \\

    Original Mean
    & 7.8761 & 7.8912 & 7.8660
    & 7.7910 & 7.7639 & 8.0811 & 7.7023 & 8.0784
    & 7.5786 & 7.9303
    & 7.9871 & 7.6137
    & 8.1084 & 8.0625
    & 7.7494 & 7.6712
    & 8.0726 & 8.0822 \\

    \lith Mean
    & \textbf{8.2924} & \textbf{8.2662} & \textbf{8.3099}
    & \textbf{8.2390} & \textbf{8.2757} & \textbf{8.3752} & \textbf{8.1876} & \textbf{8.3847}
    & \textbf{8.0271} & \textbf{8.3802}
    & \textbf{8.3456} & \textbf{8.2291}
    & \textbf{8.3640} & \textbf{8.3826}
    & \textbf{8.2299} & \textbf{8.1595}
    & \textbf{8.3644} & \textbf{8.3982} \\
    \bottomrule
  \end{tabular}
  }
\end{table*}

\section{Experimental Details}
\subsection{Experimental Settings}
\label{sec:ex_settings}

To validate the effectiveness of the \lith benchmark, we systematically evaluate multiple closed-source and open-source large vision-language models under a unified inference setting. The evaluated models include Claude-Sonnet-4.6~\cite{anthropic2026sonnet46}, Gemini-3-Pro~\cite{googledeepmind2025gemini3pro}, GPT-5.4~\cite{openai2026gpt54}, Grok-4~\cite{xai2025grok4}, Qwen3.5-27B~\cite{bai2025qwen3}, Qwen3.5-397B-A17B~\cite{bai2025qwen3}, Intern-S1-Pro~\cite{zou2026intern}, InternVL3.5-241B-A28B~\cite{wang2025internvl35}, Kimi-K2.5~\cite{moonshot2026kimik25}, and GLM-4.6V~\cite{zai2025glm46v}. These models cover different sources, parameter scales, and architectural types, providing a broad comparison of current mainstream large vision-language models on remote sensing lithology interpretation tasks. In addition, we conduct LoRA fine-tuning on Qwen3.5-9B~\cite{bai2025qwen3} and InternVL3.5-8B-HF~\cite{wang2025internvl35} using the \lith training data, with each model trained for 3 epochs, to examine the effect of domain-specific supervision on remote sensing lithology interpretation ability~\cite{hu2022lora}.

All fine-tuning experiments are conducted in a Windows environment with an NVIDIA A800 GPU and an Intel(R) Xeon(R) Silver 4210R CPU @ 2.40GHz. The two LoRA training experiments use the same training configuration: batch size $=20$, learning rate $=1\times10^{-5}$, LoRA rank $=16$, and training epochs $=3$. The training set contains 196,348 samples, including 78,543 MCQ samples and 117,805 OPEN samples. To ensure the reliability of the experimental conclusions, the training set and the final benchmark set are strictly separated at the image level, with no identical image appearing in both the training and benchmark sets. Therefore, no image-level overlap exists between the training and benchmark sets, which reduces the risk of direct data leakage.

In our LoRA fine-tuning setting, we adopt a parameter-efficient strategy rather than updating the full model parameters. The training script automatically searches for eligible \texttt{Linear} layers and injects LoRA adapters into the language-model backbone modules. Under the current training command, neither \texttt{--train\_vision} nor \texttt{--train\_projector} is enabled. Therefore, the vision encoder and the multimodal projector/merger are excluded from LoRA fine-tuning. The \texttt{lm\_head} and embedding layers are also excluded. As a result, LoRA mainly updates the \texttt{Linear} layers inside the LLM backbone, including the attention projection layers \texttt{q\_proj}, \texttt{k\_proj}, \texttt{v\_proj}, and \texttt{o\_proj}, as well as the feed-forward network layers \texttt{gate\_proj}, \texttt{up\_proj}, and \texttt{down\_proj}. The original model parameters remain frozen, and only the low-rank LoRA parameters are optimized during training.

In the inference evaluation stage, all models use the same decoding parameters: \texttt{MAX\_TOKENS} $=1024$ and \texttt{TEMPERATURE} $=0.0$. This unified inference configuration controls additional fluctuations caused by different sampling strategies and makes the horizontal comparison across models more stable and interpretable.


\subsection{Detailed Evaluation Metrics}
\label{sec:d_eva_metrics}

\lith contains both multiple-choice questions (MCQs) and open-ended questions (OPEN), so we use question-type-specific evaluation metrics. For MCQs, evaluation is based on exact option matching between the parsed model output and the ground-truth option, with accuracy as the core metric. This metric directly reflects model performance in target recognition, diagnostic feature discrimination, and objective judgment. If the model output cannot be parsed into a valid option, it is treated as an incorrect prediction.

For OPEN questions, we report two scores, OPEN$_{\mathrm{w/oJ}}$ and OPEN$_{\mathrm{J}}$, as defined in Sec.~\ref{sec:experiment_metrics}. OPEN$_{\mathrm{w/oJ}}$ is based on reference-based automatic metrics without LLM evaluation, combining ROUGE-L~\cite{lin2004rouge} and BLEU-4~\cite{papineni2002bleu}. BLEU-4 measures phrase-level $n$-gram matching, while ROUGE-L measures longest-common-subsequence overlap and better captures global structure consistency. The non-judge score assigns a larger weight to ROUGE-L because lithology interpretation often permits synonymous rewriting, varied sentence structures, and multiple valid expression paths. ROUGE-L therefore better reflects key semantic and structural consistency, while BLEU-4 provides complementary phrase-level precision and reduces overly loose matching.

OPEN$_{\mathrm{J}}$ further incorporates LLM-as-a-Judge evaluation. Automatic metrics are retained to preserve the objectivity and reproducibility of reference-based evaluation, while the judge score captures geological correctness, visual-grounding consistency, answer completeness, and professional expression quality beyond lexical overlap. For all OPEN evaluations, GPT-5.4 is used as the judge model, and the same scoring prompt is applied to all evaluated models to ensure consistency.

For each sample, the judge model is given the question, the reference answer, and the model-generated answer. It assigns four scores on a 0--10 scale: correctness, completeness, faithfulness, and overall quality. Correctness measures whether the answer is factually and geologically consistent with the reference answer; completeness reflects whether the response covers the key conclusion, relevant evidence, and necessary reasoning; faithfulness evaluates whether the answer remains grounded in the given image/question context and avoids unsupported claims; and the overall score summarizes the final quality of the response. As shown in Fig.~\ref{fig:prompt_llm_judge_evaluation}, the prompt enforces a strict JSON output format, which makes the judge results easy to parse and combine with automatic metrics.

\begin{promptbox}{Prompt for LLM-as-a-Judge Evaluation}
You are an expert evaluator for an image-based geology question-answering benchmark.

Your task is to evaluate the model answer by comparing it with the reference answer and the question.
The evaluation should focus on geological correctness, visual-grounding consistency, and answer quality.

Question:
{question}

Reference Answer:
{standard_answer}

Model Answer:
{model_answer}

Please score the model answer on a 0-10 scale for the following four dimensions:

1. correctness_score:
   - Measures whether the model answer is factually and geologically correct.
   - Give a high score if the lithology, geological phenomenon, diagnostic features, or reasoning are consistent with the reference answer.
   - Minor wording differences or synonymous geological terms should not be penalized.
   - Penalize incorrect lithology identification, wrong geological interpretation, wrong causal explanation, or conclusions that contradict the reference answer.
   - For multiple-choice questions, correctness should mainly depend on whether the selected option matches the ground-truth answer.

2. completeness_score:
   - Measures whether the model answer covers the key information expected by the reference answer.
   - Give a high score if the answer includes the main conclusion, relevant visual/geological evidence, and necessary reasoning.
   - Penalize answers that are correct but too brief, vague, or missing important diagnostic details.
   - Do not require the answer to be identical to the reference answer, but it should cover the essential points.

3. faithfulness_score:
   - Measures whether the model answer is faithful to the image/question/reference context.
   - Give a high score if the answer only uses evidence that is supported by the image or the question.
   - Penalize hallucinated details, unsupported observations, invented geological structures, or claims not grounded in the image.
   - Penalize answers that introduce irrelevant information or over-interpret the image beyond available evidence.
   - If the answer is correct but includes unsupported extra claims, reduce this score.

4. overall_score:
   - Provides a holistic score for the model answer.
   - It should reflect the combined quality of correctness, completeness, and faithfulness.
   - Correctness is the most important factor, followed by completeness and faithfulness.
   - A generally useful answer with minor omissions can receive 7-8.
   - A partially correct answer with missing evidence or some uncertainty should receive 4-6.
   - An incorrect, irrelevant, hallucinated, or empty answer should receive 0-3.
   - The overall_score should usually not be much higher than the correctness_score.

General scoring guideline:
- 9-10: Excellent answer; correct, complete, faithful, and well-supported.
- 7-8: Mostly correct; minor omissions or minor imprecision.
- 5-6: Partially correct; contains some useful information but misses key points.
- 3-4: Weak answer; limited correctness or significant omissions.
- 1-2: Mostly incorrect or largely unsupported.
- 0: Empty, irrelevant, completely incorrect, or impossible to evaluate.

Important rules:
- Do not be overly strict about wording differences.
- Accept semantically equivalent geological descriptions.
- Do not reward long answers if they contain hallucinated or unsupported information.
- Do not use information outside the question, reference answer, and model answer.
- Return strictly valid JSON only.
- Do not include explanations, comments, or markdown.

Return format:
{
  "correctness_score": 0,
  "completeness_score": 0,
  "faithfulness_score": 0,
  "overall_score": 0
}
\end{promptbox}

\captionof{figure}{Prompt for LLM-as-a-Judge evaluation of open-ended model responses.}
\label{fig:prompt_llm_judge_evaluation}

\
\subsection{Effect of Domain-Specific Fine-Tuning}
We also examine the effect of domain-specific supervision by comparing the original and LoRA fine-tuned versions of Qwen3.5-9B and InternVL3.5-8B-HF. After fine-tuning on the \lith training data, both models show clear improvements on multiple-choice accuracy and open-ended evaluation scores. This indicates that even relatively small LVLMs can better capture lithology-specific visual cues, terminology, and reasoning patterns after exposure to domain-specific supervision.

The improvement is especially evident for multiple-choice questions, where the model needs to distinguish visually similar lithological categories. For open-ended questions, fine-tuning also improves the completeness and professional consistency of generated explanations. These results support the value of \lith not only as an evaluation benchmark, but also as a training resource for developing geological knowledge-guided multimodal models.

\subsection{Additional Qualitative Observations}
\label{sec:additional_qualitative_observations}

We further provide qualitative examples to better understand the strengths and failure patterns of different LVLMs on \lith. 
As shown in Figs.~\ref{fig:ID_open_response_comparison}--\ref{fig:Reasoning_open_response_comparison}, the fine-tuned models generally produce more image-grounded and geologically consistent responses across different capability levels. 
For Identification and Description, the fine-tuned InternVL model accurately captures the gray-white tone, coarse mottled texture, weak structural fabric, and massive intrusive character of the diorite example, while some stronger general models provide broader but less specific alternatives such as granite, granodiorite, or metamorphic rock. 
For Comparative Analysis, Mechanism Explanation, Practical Application, and Comprehensive Reasoning questions, fine-tuned models more consistently connect visual cues with geological interpretation, such as using warm earthy tones to distinguish arenaceous rocks from crystalline surfaces, relating coarse texture and weak layering to granite or diorite formation, and assessing quarry suitability from weathering, blocky fragmentation, and joint-related concerns. 
In contrast, several non-fine-tuned models tend to produce plausible but less reliable answers, including over-generalized descriptions, unsupported biological or crystalline interpretations, and incomplete links between visible evidence and geological mechanisms. 
The MCQ examples in Fig.~\ref{fig:mcq_ft_case_comparison} further support this observation: in challenging cases with visually similar distractors, both fine-tuned models select the correct option, whereas the corresponding base models and many other models are misled by plausible alternatives. 
These results suggest that the difficulty of \lith does not only lie in recognizing image patterns, but also in grounding visual evidence in domain-specific lithological knowledge. 
They also indicate that domain fine-tuning improves both objective choice accuracy and open-ended geological reasoning quality.


\begin{figure}[H]
\centering
\begin{casebox}{Identification and Description of Open-ended Lithology Responses}
\scriptsize

\noindent
\begin{minipage}[t]{0.15\linewidth}
  \vspace{0pt}
  \centering
  \includegraphics[width=\linewidth]{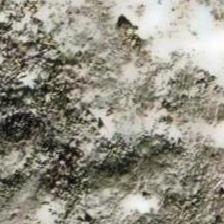}
  \vspace{1mm}
  
\end{minipage}
\hspace{3mm}
\begin{minipage}[t]{0.8\linewidth}
  \vspace{0pt}
  \small
  \textbf{Question.}
  Based on the visible tone, texture, and structural expression in this image, how would you identify and describe the lithology present here? Please comment on its color/brightness, grain character, degree of mottling or speckling, continuity of the exposed surface, and whether any clear bedding or strong linear structures are present.
\end{minipage}

{\fontsize{8pt}{10pt}\selectfont
\textbf{[Reference Answer]}\\[1mm]
The image is most consistent with diorite. It shows a medium overall brightness with a dominant gray to gray-white color balance and moderate contrast between lighter exposed patches and darker mottled areas. The surface texture appears coarse and distinctly heterogeneous, with abundant dark flecks and granular mottling that give it a moderate to strong speckled appearance. The exposed rock forms a largely continuous surface rather than isolated fragments, although tonal domains are irregular and somewhat blocky in outline. Structural organization is weak: there is no obvious bedding, no persistent banding, and no strong linear fabric across the scene. Hydro-geomorphic overprint is also limited, with low visible gully development and only weak diffuse roughness. Overall, this is a massive, coarse-grained intrusive rock surface with the mottled gray appearance typical of diorite.

\vspace{2mm}

\noindent
\begin{tabular*}{\linewidth}{@{\extracolsep{\fill}}lr@{}}
\textbf{[InternVL3.5-8B-HF(FT)]} &
{\footnotesize
\textit{B-4}: \textbf{0.2217} \;|\;
\textit{R-L}: \textbf{0.5018} \;|\;
\textit{JudgeOverall}: \textbf{10.0000} \;|\;
Open$_{\mathrm{J}}$: \textbf{6.4505}
}
\end{tabular*}\\[1mm]
The image is consistent with diorite. It shows a medium overall brightness with a clear gray to slightly gray-green hue bias. Contrast is moderate, produced by scattered darker mottled patches against lighter exposed areas. The texture is distinctly coarse and heterogeneous, with strong speckling and irregular mottling rather than a smooth or uniform surface. The rock appears as a broad, continuous exposure rather than isolated fragments, and internal boundaries are diffuse. Structural organization is weak: there are no obvious continuous bedding planes, no strong banding, and no prominent linear features such as folds or fault traces. Any tonal alignment is faint and discontinuous. Hydro-geomorphic expression is also subdued, with low gully density and no clear fan-shaped deposits or slump features. Overall, the scene is best described as a massive, coarse-grained, speckled intrusive rock surface with diffuse internal variation and weak structural fabric.

\vspace{2mm}

\noindent
\begin{tabular*}{\linewidth}{@{\extracolsep{\fill}}lr@{}}
\textbf{[GPT-5.4]} &
{\footnotesize
\textit{B-4}: \textbf{0.0332} \;|\;
\textit{R-L}: \textbf{0.2000} \;|\;
\textit{JudgeOverall}: \textbf{7.0000} \;|\;
Open$_{\mathrm{J}}$: \textbf{3.6665}
}
\end{tabular*}\\[1mm]
The exposed material appears to be a **light-colored, massive crystalline rock**, most consistent in appearance with a **felsic intrusive lithology such as granite or granitic gneiss**, though the image alone does not allow a firm identification. Its overall **color/brightness is pale gray to whitish**, with strong **dark gray to black mottling/speckling** distributed irregularly across the surface. That speckled character suggests a **medium- to coarse-grained crystalline texture**, where darker mafic minerals are scattered through a lighter feldspar- and quartz-rich groundmass.The surface exposure looks fairly **continuous and coherent**, not obviously crumbly or bedded, which supports interpretation as a competent crystalline rock rather than a layered sedimentary unit. The mottling is patchy and diffuse rather than forming regular laminae. I do **not see clear bedding planes** or any persistent, strong linear structures; at most there are weak, irregular streaks or tonal variations that could reflect weathering, mineral segregation, or faint foliation. Overall, this looks like a **massive, speckled crystalline lithology** with little obvious stratification.

\vspace{2mm}

\noindent
\begin{tabular*}{\linewidth}{@{\extracolsep{\fill}}lr@{}}
\textbf{[Qwen3.5-397B-A17B]} &
{\footnotesize
\textit{B-4}: \textbf{0.0370} \;|\;
\textit{R-L}: \textbf{0.1818} \;|\;
\textit{JudgeOverall}: \textbf{8.0000} \;|\;
Open$_{\mathrm{J}}$: \textbf{4.0013}
}
\end{tabular*}\\[1mm]
Based on the visual evidence, the lithology appears to be a coarse-grained, crystalline rock, likely an intrusive igneous rock such as diorite or granodiorite, or potentially a weakly foliated metamorphic rock. The color and brightness are highly variegated, featuring a high-contrast mix of bright white to pale grey areas (likely feldspar or quartz) interspersed with dark green to black patches (possibly amphibole, biotite, or pyroxene). The grain character is phaneritic, meaning individual mineral crystals are distinct and interlocking, giving the surface a rough, granular texture. There is a significant degree of mottling and speckling, creating a classic \"salt-and-pepper\" appearance. The exposed surface appears massive and continuous, lacking the smooth, uniform finish typical of fine-grained sedimentary stones. Regarding structure, there are no clear sedimentary bedding planes visible. While there is a slight, chaotic alignment in some darker mineral clusters, there are no strong, continuous linear structures or distinct banding, suggesting a massive rock fabric rather than a strongly layered one.
}

\end{casebox}

\caption{Qualitative comparison of open-ended lithology responses for the Identification and Description task.}
\label{fig:ID_open_response_comparison}
\end{figure}


\begin{figure}[H]
\centering
\begin{casebox}{Comparative Analysis of Open-ended Lithology Responses}
\scriptsize

\noindent
\begin{minipage}[t]{0.15\linewidth}
  \vspace{0pt}
  \centering
  \includegraphics[width=\linewidth]{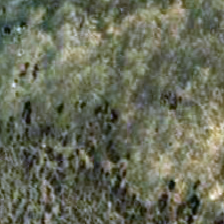}
  \vspace{1mm}
  
\end{minipage}
\hspace{3mm}
\begin{minipage}[t]{0.8\linewidth}
  \vspace{0pt}
  \small
  \textbf{Question.}
What visual characteristics in this image would help an expert distinguish it from a similarly mottled crystalline-looking surface, especially in terms of color tone, texture, and structural organization?
\end{minipage}

{\fontsize{8pt}{10pt}\selectfont
\textbf{[Reference Answer]}\\[1mm]
The most useful distinguishing cues are its warmer yellow-brown to olive tone, medium rather than coarse granularity, and its less blocky, more diffusely mottled organization. The surface shows moderate speckling with irregular dark patches dispersed through a lighter earthy background, and some of these darker areas loosely link into weak dendritic or reticulate patterns. Any linear alignment or bedding is only faintly expressed, but the overall appearance suggests a patchy, somewhat heterogeneous sedimentary surface rather than a rougher, grayer, more crystalline-looking texture. Boundaries between tonal domains are soft and gradual, and erosion appears as diffuse slope wash with only shallow incision, reinforcing a broadly continuous arenaceous unit rather than a more coarse, blocky massive rock surface.

\vspace{2mm}

\noindent
\begin{tabular*}{\linewidth}{@{\extracolsep{\fill}}lr@{}}
\textbf{[Qwen3.5-9B(FT)]} &
{\footnotesize
\textit{B-4}: \textbf{0.0930} \;|\;
\textit{R-L}: \textbf{0.2510} \;|\;
\textit{JudgeOverall}: \textbf{9.0000} \;|\;
Open$_{\mathrm{J}}$: \textbf{4.7900}
}
\end{tabular*}\\[1mm]
The most useful distinguishing traits are the warmer yellow-brown to olive-brown tone and the more diffuse, earthy surface character. Although both surfaces are mottled, medium-grained, and weakly structured, this image lacks the cooler blue-gray cast typical of a crystalline-looking surface. Its texture is moderately speckled but not strongly blocky, and the internal boundaries are soft and diffuse rather than sharply defined. Structural organization is weak in both cases, with little banding or linear fabric, but here the surface reads more as a continuous sedimentary exposure with subtle mottling and faint erosional streaking. There are no strong arcuate forms, dense gullies, or obvious blocky fragmentation. Overall, the combination of warm earthy coloration, moderate mottling, and diffuse internal variation is the key visual cue separating it from a more crystalline, cool-toned rock surface.

\vspace{2mm}

\noindent
\begin{tabular*}{\linewidth}{@{\extracolsep{\fill}}lr@{}}
\textbf{[Qwen3.5-9B]} &
{\footnotesize
\textit{B-4}: \textbf{0.0148} \;|\;
\textit{R-L}: \textbf{0.1087} \;|\;
\textit{JudgeOverall}: \textbf{5.0000} \;|\;
Open$_{\mathrm{J}}$: \textbf{2.4643}
}
\end{tabular*}\\[1mm]
The primary distinction lies in the organic, amorphous nature of the surface. Unlike crystalline structures which typically exhibit sharp geometric facets, high reflectivity, and distinct angular boundaries, this image displays a soft, matte texture. The color palette is dominated by muted, earthy tones—specifically a blend of olive greens, dull browns, and grayish-beiges—which suggests biological material like moss, lichen, or perhaps a close-up of skin or leaf tissue. The structural organization is irregular and fluid; the \"mottling\" consists of soft-edged, blended patches rather than the rigid, repeating lattice patterns or distinct grain boundaries found in minerals. The overall impression is one of depth and fuzziness, lacking the hard, crystalline sheen and precise symmetry associated with geological crystals.

\vspace{2mm}

\noindent
\begin{tabular*}{\linewidth}{@{\extracolsep{\fill}}lr@{}}
\textbf{[Intern-S1-Pro]} &
{\footnotesize
\textit{B-4}: \textbf{0.0215} \;|\;
\textit{R-L}: \textbf{0.1169} \;|\;
\textit{JudgeOverall}: \textbf{4.0000} \;|\;
Open$_{\mathrm{J}}$: \textbf{2.1107}
}
\end{tabular*}\\[1mm]
To distinguish this image from a mottled crystalline surface, an expert would focus on three key aspects: **Color tone**: The image exhibits *earthy, muted organic hues* (greens, browns, and yellowish tints) with no metallic sheen or high-contrast reflective highlights typical of crystalline structures. Crystalline surfaces often display sharp, uniform color gradients or iridescence due to light refraction, whereas this image’s coloration is diffuse and non-reflective, suggesting biological or weathered geological material. **Texture**: The surface shows *irregular, granular roughness* with no defined facets, edges, or smooth cleavage planes characteristic of crystals. Crystalline textures typically feature geometric precision (e.g., hexagonal or cubic patterns), but this image’s texture appears chaotic, porous, and uneven—reminiscent of lichen, moss, or eroded rock. **Structural organization**: There is *no periodic or symmetrical patterning*; the “mottling” is random and lacks the repeating unit cells of crystalline lattices. Crystalline surfaces often display ordered microstructures (e.g., dendritic or faceted growth), while this image’s irregular dark spots and amorphous patches suggest organic growth or stochastic mineral deposition, not atomic-scale order. In summary, the absence of crystalline symmetry, non-reflective organic coloration, and chaotic texture would signal this is not a crystalline surface but rather a biological or weathered geological specimen.

}

\end{casebox}

\caption{Qualitative comparison of open-ended lithology responses for the Comparative Analysis task.}
\label{fig:Comparison_open_response_comparison}
\end{figure}


\begin{figure}[H]
\centering
\begin{casebox}{Mechanism Explanation of Open-ended Lithology Responses}
\scriptsize

\noindent
\begin{minipage}[t]{0.15\linewidth}
  \vspace{0pt}
  \centering
  \includegraphics[width=\linewidth]{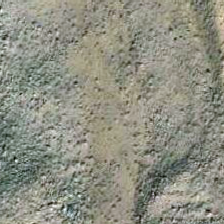}
  \vspace{1mm}
  
\end{minipage}
\hspace{3mm}
\begin{minipage}[t]{0.8\linewidth}
  \vspace{0pt}
  \small
  \textbf{Question.}
Why is this image more consistent with a granite exposure than with a layered sedimentary or foliated metamorphic rock? Explain using the observed tone, coarse texture, structural pattern, and what these features imply about mineral composition and crystallization.

\end{minipage}

{\fontsize{8pt}{10pt}\selectfont
\textbf{[Reference Answer]}

This scene is most consistent with granite because it shows a bright, pale beige to light gray surface with a coarse, mottled granular texture and no clear bedding or foliation. The light overall tone fits a felsic composition dominated by quartz and feldspar, which commonly gives granite a pale appearance in remote-sensing imagery. The moderate speckling and uneven patchiness suggest visible variation among mineral grains or small mineral aggregates rather than sedimentary layering. Just as importantly, the image lacks persistent parallel bands, repeated stratification, or a strong planar fabric, so it does not resemble a bedded sedimentary unit or a strongly foliated metamorphic rock. The weak linear traces and scattered darker patches are better interpreted as localized fractures, weathered zones, or mafic mineral concentrations within an otherwise massive crystalline rock. Altogether, the broad continuous exposure, coarse crystalline appearance, and absence of organized layering support an intrusive igneous rock that cooled slowly enough to develop coarse crystals, which is characteristic of granite.

\vspace{2mm}

\noindent
\begin{tabular*}{\linewidth}{@{\extracolsep{\fill}}lr@{}}
\textbf{[Qwen3.5-9B(FT)]} &
{\footnotesize
\textit{B-4}: \textbf{0.1169} \;|\;
\textit{R-L}: \textbf{0.2710} \;|\;
\textit{JudgeOverall}: \textbf{9.0000} \;|\;
Open$_{\mathrm{J}}$: \textbf{4.9177}
}
\end{tabular*}\\[1mm]
This scene is most consistent with granite because it shows a bright gray to off-white tone, a coarse mottled texture, and only weak linear structure. The pale color is compatible with abundant felsic minerals such as quartz and feldspar, which commonly produce light reflectance in remote-sensing imagery. The coarse granularity suggests slow crystallization at depth, allowing interlocking mineral grains to become visible rather than forming fine sedimentary layers. The image lacks clear bedding, rhythmic banding, or strong foliation, so it does not resemble stratified sedimentary rocks or strongly deformed metamorphic rocks. Instead, the surface is broadly massive and moderately homogeneous, with diffuse internal tonal variation rather than organized planar structures. The faint darker streaks are better interpreted as minor compositional or weathering variations within a felsic crystalline rock mass. Overall, the combination of light tone, coarse crystalline texture, and absence of sedimentary or metamorphic fabric supports identification as granite.

\vspace{2mm}

\noindent
\begin{tabular*}{\linewidth}{@{\extracolsep{\fill}}lr@{}}
\textbf{[Grok-4]} &
{\footnotesize
\textit{B-4}: \textbf{0.0685} \;|\;
\textit{R-L}: \textbf{0.1662} \;|\;
\textit{JudgeOverall}: \textbf{9.0000} \;|\;
Open$_{\mathrm{J}}$: \textbf{4.4018}
}
\end{tabular*}\\[1mm]
The image shows a surface with a **coarse, speckled texture** and **overall light-to-medium grayish tone** featuring irregular patches of slightly darker and lighter areas. This is far more consistent with a **granite exposure** than with layered sedimentary or foliated metamorphic rock. Granite is an **intrusive igneous rock** formed by slow cooling of magma deep underground, producing large, interlocking crystals of quartz (light), feldspar (pinkish-white to gray), and biotite or hornblende (dark specks). The observed **coarse, granular texture** reflects this phaneritic (visible-crystal) structure, with no preferred alignment or banding. In contrast, **layered sedimentary rocks** typically display **stratification**—distinct horizontal bedding, laminations, or color banding from deposition over time—which is absent here. **Foliated metamorphic rocks** (e.g., gneiss or schist) show **parallel mineral alignment** or wavy banding due to directed pressure and recrystallization, creating a linear or planar fabric. The image’s **randomly distributed speckles** and **lack of any linear or planar structural pattern** indicate isotropic crystallization typical of granite, not the anisotropic textures of sedimentary layering or metamorphic foliation. The mottled tone suggests mixed felsic minerals without segregation into layers.

\vspace{2mm}

\noindent
\begin{tabular*}{\linewidth}{@{\extracolsep{\fill}}lr@{}}
\textbf{[Kimi-K2.5]} &
{\footnotesize
\textit{B-4}: \textbf{0.0780} \;|\;
\textit{R-L}: \textbf{0.1911} \;|\;
\textit{JudgeOverall}: \textbf{9.0000} \;|\;
Open$_{\mathrm{J}}$: \textbf{4.5203}
}
\end{tabular*}\\[1mm]
The image displays a **massive, coarse-grained rock** with a **mottled, speckled tone** of intermixed light and dark minerals, lacking any visible layering or preferred orientation. This texture is diagnostic of **granite**, an intrusive igneous rock. The **coarse, interlocking crystalline texture** indicates slow cooling deep underground, allowing large mineral crystals to form. The random distribution of light-colored quartz and feldspar with darker biotite or hornblende creates the salt-and-pepper appearance. In contrast, **layered sedimentary rocks** would exhibit distinct bedding planes and grain sorting, while **foliated metamorphic rocks** would show aligned minerals or compositional banding from directed pressure. The absence of stratification or foliation, combined with the massive, crystalline structure, confirms this rock formed through **crystallization from magma** rather than deposition in layers or recrystallization under stress. The interlocking crystal boundaries visible in the texture are characteristic of igneous rocks that cooled slowly enough for minerals to grow into each other.

}

\end{casebox}

\caption{Qualitative comparison of open-ended lithology responses for the Mechanism Explanation task.}
\label{fig:Explanation_open_response_comparison}
\end{figure}


\begin{figure}[H]
\centering
\begin{casebox}{Practical Application of Open-ended Lithology Responses}
\scriptsize

\noindent
\begin{minipage}[t]{0.15\linewidth}
  \vspace{0pt}
  \centering
  \includegraphics[width=\linewidth]{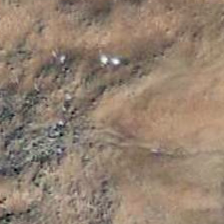}
  \vspace{1mm}
  
\end{minipage}
\hspace{3mm}
\begin{minipage}[t]{0.8\linewidth}
  \vspace{0pt}
  \small
  \textbf{Question.}
Based on the exposed surface texture, moderate blocky fragmentation, and weak linear structure visible in this diorite outcrop, how suitable would this area be for quarrying high-quality decorative facing stone, and what field checks would you prioritize before any extraction decision?

\end{minipage}

{\fontsize{8pt}{10pt}\selectfont
\textbf{[Reference Answer]}\\[1mm]
This exposure appears only moderately suitable, and probably not ideal, for quarrying premium decorative facing stone. The image shows broad exposed rock, which is favorable for access to continuous material, but the surface is texturally heterogeneous, with rough mottled zones on the left, smoother zones on the right, diffuse boundaries, and moderate blocky fragmentation. For decorative stone, industry preference is usually for massive, visually uniform rock with limited fracturing so that large slabs can be extracted and polished consistently. Here, the weak but present streaking and variable weathered expression suggest that appearance and block integrity may change across short distances. The yellow-brown tone also looks strongly weathered, so the visible surface may not represent the true fresh-rock color or fabric. Before any quarry decision, I would prioritize checking fresh broken surfaces, mapping joint spacing and persistence, measuring recoverable block size, and testing color and grain uniformity on cut slabs. Trial extraction and image-based fabric assessment on fresh material would be essential before evaluating commercial value.

\vspace{2mm}

\noindent
\begin{tabular*}{\linewidth}{@{\extracolsep{\fill}}lr@{}}
\textbf{[Qwen3.5-9B(FT)]} &
{\footnotesize
\textit{B-4}: \textbf{0.0595} \;|\;
\textit{R-L}: \textbf{0.1847} \;|\;
\textit{JudgeOverall}: \textbf{9.0000} \;|\;
Open$_{\mathrm{J}}$: \textbf{4.4579}
}
\end{tabular*}\\[1mm]
The exposed diorite may have some quarry potential, but the image does not indicate an ideal source for premium decorative facing stone. The surface is weathered, with medium yellow-brown tones, moderate mottling, and moderate blocky fragmentation, suggesting limited visual uniformity and possible discontinuities. For high-quality facing slabs, operators usually prefer massive, fresh, homogeneous rock with consistent color and texture. The weak linear features and patchy darker zones also raise concern about reduced block recovery and variable appearance. Before extraction, field teams should prioritize inspection of fresh fractured surfaces rather than the weathered exterior, because weathering can obscure true mineral texture and color. They should also map joint spacing, assess recoverable block size, and test durability and polish response. In short, the site may be usable for lower-grade stone or crushed aggregate, but it is not yet proven to be a top-quality decorative stone quarry.

\vspace{2mm}

\noindent
\begin{tabular*}{\linewidth}{@{\extracolsep{\fill}}lr@{}}
\textbf{[Claude-Sonnet-4.6]} &
{\footnotesize
\textit{B-4}: \textbf{0.0391} \;|\;
\textit{R-L}: \textbf{0.1702} \;|\;
\textit{JudgeOverall}: \textbf{9.0000} \;|\;
Open$_{\mathrm{J}}$: \textbf{4.3591}
}
\end{tabular*}\\[1mm]
Based on the visible characteristics, this diorite outcrop shows **moderate suitability** for decorative stone, but requires careful evaluation. **Concerns:**- The blocky fragmentation and weathered surface suggest potential joint sets that could limit extractable block size.- Weak linear structure may indicate fracturing that compromises integrity- Surface weathering obscures fresh rock quality assessment **Priority field checks:**1. **Joint spacing and orientation** - Map discontinuities to determine maximum recoverable block dimensions. 2. **Fresh rock exposure** - Break samples to assess unweathered color uniformity, grain size consistency, and aesthetic appeal. 3. **Soundness testing** - Schmidt hammer tests for surface hardness; check for incipient fractures. 4. **Depth assessment** - Excavate test pits to verify if quality improves with depth and weathering decreases. 5. **Petrographic analysis** - Thin sections to evaluate mineral composition, alteration, and potential for polishing. 6. **Volume estimation** - Calculate economically viable reserve considering waste from jointing. The moderate fragmentation suggests this may be better suited for smaller architectural elements rather than large facing panels, unless subsurface conditions improve significantly.

\vspace{2mm}

\noindent
\begin{tabular*}{\linewidth}{@{\extracolsep{\fill}}lr@{}}
\textbf{[InternVL3.5-241B-A28B]} &
{\footnotesize
\textit{B-4}: \textbf{0.0330} \;|\;
\textit{R-L}: \textbf{0.1786} \;|\;
\textit{JudgeOverall}: \textbf{6.0000} \;|\;
Open$_{\mathrm{J}}$: \textbf{3.1802}
}
\end{tabular*}\\[1mm]
The diorite outcrop's suitability for high-quality decorative facing stone depends on several factors observed in the image. Diorite’s inherent properties—medium-coarse grain, durability, and aesthetic speckled appearance—make it a candidate for decorative use. However, the **moderate blocky fragmentation** suggests favorable jointing for extracting rectangular blocks, though excessive fracturing could limit yields of large, intact slabs. The **weak linear structures** may indicate subtle foliation or incipient jointing; if these are sparse and non-persistent, they might not compromise stone quality, but pervasive alignment could lead to uneven breaking or anisotropic strength. The **exposed surface texture** appears relatively fresh, but any weathering (e.g., surface rind, discoloration) would reduce polishability and visual appeal. **Prioritized field checks** include:  1. **Joint Mapping**: Assess spacing, orientation, and persistence of fractures to predict block size and extraction feasibility.  2. **Core Sampling**: Evaluate internal homogeneity, mineralogy, and hidden defects (e.g., alteration zones, veins).  3. **Durability Tests**: Measure compressive strength, porosity, and weathering resistance to ensure longevity.  4. **Aesthetic Assessment**: Confirm consistent color, grain size, and absence of disruptive mineral inclusions (e.g., oxidized pyrite).  5. **Permeability Checks**: Low permeability is critical for a smooth polish.  If fragmentation is manageable and linear features are minimal, the site could yield premium stone. However, if weaknesses are widespread, the material may be better suited for aggregate than decorative use.
}

\end{casebox}

\caption{Qualitative comparison of open-ended lithology responses for the Practical Application task.}
\label{fig:Application_open_response_comparison}
\end{figure}


\begin{figure}[H]
\centering
\begin{casebox}{Comprehensive Reasoning of Open-ended Lithology Responses}
\scriptsize

\noindent
\begin{minipage}[t]{0.15\linewidth}
  \vspace{0pt}
  \centering
  \includegraphics[width=\linewidth]{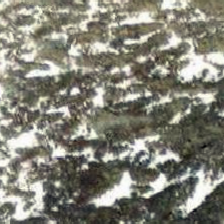}
  \vspace{1mm}
  
\end{minipage}
\hspace{3mm}
\begin{minipage}[t]{0.8\linewidth}
  \vspace{0pt}
  \small
  \textbf{Question.}
Given the image’s coarse mottled texture, strong bright–dark mineral contrast, and the lack of clear bedding, foliation, or cross-cutting linear structures, what geological origin is most reasonably inferred for this rock mass, and how do the visible features support that interpretation?

\end{minipage}

{\fontsize{8pt}{10pt}\selectfont
\textbf{[Reference Answer]}\\[1mm]
The most reasonable interpretation is that this is a massive intrusive igneous rock, specifically a plutonic body such as diorite formed by slow cooling at depth. The key reasoning is the combination of coarse granularity and low homogeneity: individual light and dark mineral patches are large enough to be visually distinguished, which indicates relatively slow crystallization rather than rapid cooling. The image also lacks planar sedimentary layering, repeated banding, or any consistent foliation that would suggest deposition or strong metamorphic fabric development. Instead, the rock surface appears as an irregular interlocking mosaic of bright and dark components, consistent with contrasting felsic and mafic minerals crystallized together in a crystalline intrusive rock. The broadly continuous exposure and only weak linear structure further support a massive rock body rather than stratified units. Therefore, the visible evidence points to underground magma crystallization producing a coarse-grained dioritic texture rather than a sedimentary or strongly deformed metamorphic origin.

\vspace{2mm}

\noindent
\begin{tabular*}{\linewidth}{@{\extracolsep{\fill}}lr@{}}
\textbf{[InternVL3.5-8B-HF(FT)]} &
{\footnotesize
\textit{B-4}: \textbf{0.1508} \;|\;
\textit{R-L}: \textbf{0.2834} \;|\;
\textit{JudgeOverall}: \textbf{10.0000} \;|\;
Open$_{\mathrm{J}}$: \textbf{5.4350}
}
\end{tabular*}\\[1mm]
The most reasonable interpretation is that this is a massive intrusive igneous rock, specifically a plutonic rock such as diorite. The key evidence is the coarse, heterogeneous texture with irregular light and dark patches rather than layered or banded fabric. The strong tonal contrast suggests intergrown minerals of different compositions, which is consistent with crystallization of magma at depth where crystals had time to grow to visible size. The absence of persistent planar structures means there is no clear sedimentary bedding or metamorphic foliation to indicate deposition in layers or strong directed strain. Likewise, no obvious cross-cutting veins or dikes are visible, so there is no evidence for younger intrusive or brittle structures disrupting the main rock mass. Overall, the image shows a relatively homogeneous but mottled crystalline surface distributed across a continuous exposure, which supports slow cooling and crystallization underground rather than rapid surface deposition or strong tectonic deformation.Explanation: The question is explanation-oriented and tied directly to visible evidence in the target image: coarse granularity, low homogeneity, strong contrast, weak banding, weak linear structure, and large continuous exposure. The answer uses geological reasoning consistent with the provided knowledge that coarse crystalline igneous rocks form by slow underground cooling, while avoiding unsupported claims about exact mineralogy, age, or tectonic setting.

\vspace{2mm}

\noindent
\begin{tabular*}{\linewidth}{@{\extracolsep{\fill}}lr@{}}
\textbf{[InternVL3.5-8B-HF]} &
{\footnotesize
\textit{B-4}: \textbf{0.0282} \;|\;
\textit{R-L}: \textbf{0.1857} \;|\;
\textit{JudgeOverall}: \textbf{8.0000} \;|\;
Open$_{\mathrm{J}}$: \textbf{3.9991}
}
\end{tabular*}\\[1mm]
The rock mass is most reasonably inferred to originate from a plutonic igneous process, such as granite. The coarse, mottled texture and strong contrast between light and dark minerals suggest slow cooling beneath Earth's surface, allowing large crystals to form. The absence of bedding, foliation, or cross-cutting structures indicates it is not sedimentary or significantly deformed by metamorphism. These features align with intrusive igneous rocks, which crystallize from magma in a stable environment, preserving a homogeneous, interlocking mineral assemblage without tectonic disruption.

\vspace{2mm}

\noindent
\begin{tabular*}{\linewidth}{@{\extracolsep{\fill}}lr@{}}
\textbf{[GLM-4.6V]} &
{\footnotesize
\textit{B-4}: \textbf{0.0314} \;|\;
\textit{R-L}: \textbf{0.1951} \;|\;
\textit{JudgeOverall}: \textbf{9.0000} \;|\;
Open$_{\mathrm{J}}$: \textbf{4.4432}
}
\end{tabular*}\\[1mm]
The rock mass is most reasonably inferred to be an **intrusive (plutonic) igneous rock** (e.g., granite or gabbro). The coarse mottled texture and strong mineral contrast (bright vs. dark minerals) reflect slow cooling of magma, allowing large, distinct mineral grains (e.g., feldspar, quartz, mica) to form. The absence of bedding (sedimentary layering), foliation (metamorphic alignment), or cross-cutting structures (e.g., dikes) rules out sedimentary or metamorphic origins, as these features are diagnostic of those rock types. Intrusive igneous rocks form deep within the crust, where slow cooling promotes grain growth and heterogeneous mineral distribution, creating the mottled, non-layered appearance. The lack of structural fabric further supports an igneous origin, as such features arise from magma solidification rather than sedimentary deposition or tectonic deformation.
}
\end{casebox}
\caption{Qualitative comparison of open-ended lithology responses for the Comprehensive Reasoning task.}
\label{fig:Reasoning_open_response_comparison}
\end{figure}

\begin{figure}[H]
\centering
\begin{casebox}{MCQ Cases Highlighting the Effect of Domain Fine-tuning}
\scriptsize

\noindent
\begin{minipage}[t]{0.15\linewidth}
  \vspace{0pt}
  \centering
  \includegraphics[width=\linewidth]{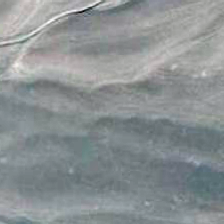}
\end{minipage}
\hspace{3mm}
\begin{minipage}[t]{0.80\linewidth}
  \vspace{0pt}
  \small
  \textbf{Case 1: Identification and Description.}\\
  \textbf{Question.}
  Based on the image's overall medium-gray tone, fine and fairly uniform texture, low speckle, and only weak diffuse banding without clearly resolved rounded or angular clasts, which lithology is most consistent with the scene?

  \vspace{1mm}
  \textbf{Options.}\\
  A. Crushed stone \quad
  B. Breccia \quad
  C. Arenaceous rocks \quad
  D. Conglomerate

  \vspace{1mm}
  \textbf{Ground truth:} \correctans{A}
\end{minipage}

\vspace{2mm}

\renewcommand{\arraystretch}{1.18}
\begin{tabular}{p{0.31\linewidth}p{0.31\linewidth}p{0.31\linewidth}}
\toprule
\modelans{Claude-Sonnet-4.6}{\wrongans{C}} &
\modelans{Gemini-3-Pro}{\wrongans{C}} &
\modelans{GPT-5.4}{\wrongans{C}} \\

\modelans{Grok-4}{\wrongans{C}} &
\modelans{Qwen3.5-27B}{\wrongans{C}} &
\modelans{Qwen3.5-397B-A17B}{\wrongans{C}} \\

\modelans{Intern-S1-Pro}{\wrongans{C}} &
\modelans{InternVL3.5-241B-A28B}{\wrongans{C}} &
\modelans{Kimi-K2.5}{\wrongans{C}} \\

\modelans{GLM-4.6V}{\wrongans{C}} &
\modelans{InternVL3.5-8B-HF}{\wrongans{C}} &
\modelans{Qwen3.5-9B}{\wrongans{C}} \\

\modelans{InternVL3.5-8B-HF(FT)}{\correctans{A}} &
\modelans{Qwen3.5-9B(FT)}{\correctans{A}} &
\textbf{Correct:} \correctans{2/14} \\
\bottomrule
\end{tabular}

\vspace{1mm}
\noindent

\vspace{4mm}

\noindent
\begin{minipage}[t]{0.15\linewidth}
  \vspace{0pt}
  \centering
  \includegraphics[width=\linewidth]{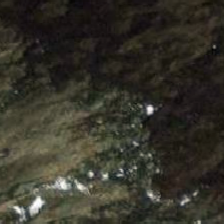}
\end{minipage}
\hspace{3mm}
\begin{minipage}[t]{0.80\linewidth}
  \vspace{0pt}
  \small
  \textbf{Case 2: Comprehensive Reasoning.}\\
  \textbf{Question.}
  The outcrop-scale surface appears as a dark, laterally continuous unit with only weak, discontinuous streaking and no strong blocky fabric or obvious cross-cutting igneous body. Based on these visible relationships, which interpretation is most reasonable for the geological character of this rock mass?

  \vspace{1mm}
  \textbf{Options.}\\
  A. A sedimentary arenaceous unit in which primary bedding is only faintly expressed and erosion has produced shallow linear wash patterns.\\
  B. A massive plutonic intrusion whose uniform texture indicates slow underground crystallization without sedimentary layering.\\
  C. A carbonate platform marked by strong karst dissolution, with circular collapse features controlling the surface pattern.\\
  D. A coarse talus or crushed-stone deposit composed of angular fragments with high local relief and sharp clast boundaries.

  \vspace{1mm}
  \textbf{Ground truth:} \correctans{A}
\end{minipage}

\vspace{2mm}

\renewcommand{\arraystretch}{1.18}
\begin{tabular}{p{0.31\linewidth}p{0.31\linewidth}p{0.31\linewidth}}
\toprule
\modelans{Claude-Sonnet-4.6}{\correctans{A}} &
\modelans{Gemini-3-Pro}{\correctans{A}} &
\modelans{GPT-5.4}{\correctans{A}} \\

\modelans{Grok-4}{\wrongans{B}} &
\modelans{Qwen3.5-27B}{\wrongans{C}} &
\modelans{Qwen3.5-397B-A17B}{\correctans{A}} \\

\modelans{Intern-S1-Pro}{\correctans{A}} &
\modelans{InternVL3.5-241B-A28B}{\wrongans{B}} &
\modelans{Kimi-K2.5}{\correctans{A}} \\

\modelans{GLM-4.6V}{\correctans{A}} &
\modelans{InternVL3.5-8B-HF}{\wrongans{B}} &
\modelans{Qwen3.5-9B}{\wrongans{B}} \\

\modelans{InternVL3.5-8B-HF(FT)}{\correctans{A}} &
\modelans{Qwen3.5-9B(FT)}{\correctans{A}} &
\textbf{Correct:} \correctans{9/14} \\
\bottomrule
\end{tabular}

\vspace{1mm}
\noindent

\end{casebox}

\caption{MCQ examples where fine-tuned LVLMs correctly answer lithology questions while base models or many other models fail. Only the predicted option key is shown for each model. Green indicates a correct answer, and red indicates an incorrect answer.}
\label{fig:mcq_ft_case_comparison}
\end{figure}

\section{Limitation}
\label{sec:limi}

Although \lith provides a dedicated benchmark for remote-sensing lithology interpretation, it still has several limitations. First, the current image data are mainly collected from GF-2 imagery over exposed surface regions in northwestern China. The visual appearance of lithological units may vary under different geological settings, climate zones, weathering conditions, vegetation coverage, illumination, and sensor characteristics. Therefore, the current benchmark may not fully represent all global lithological scenarios.

Second, \lith covers 12 representative lithology and auxiliary land-cover categories. These categories are useful for evaluating common lithology-related interpretation tasks, but they do not cover the full complexity of geological mapping. More fine-grained rock types, mineral alteration zones, structural features, and mixed lithological units remain to be incorporated in future versions.

Third, the construction pipeline uses large multimodal models and LLM-as-a-Judge modules for description generation, question-answer generation, and quality filtering. Although expert-in-the-loop design and multi-stage filtering are used to improve reliability, model-generated content may still contain biases, incomplete reasoning, or terminology inconsistencies. The benchmark should therefore be regarded as a carefully filtered evaluation resource rather than a replacement for expert geological field validation.

Fourth, open-ended evaluation remains challenging. Reference-based metrics such as BLEU-4 and ROUGE-L cannot fully capture geological correctness, while LLM-as-a-Judge evaluation may introduce judge-model preference bias. To reduce this issue, we combine automatic metrics with judge-based scoring, but future work should further explore human expert evaluation and more geology-aware evaluation protocols.

Finally, \lith focuses on patch-level interpretation rather than full-scene geological mapping. Real-world lithology mapping often requires multi-scale spatial context, stratigraphic relationships, structural geology, geochemical information, and field observations. Therefore, models evaluated on \lith should be considered as decision-support tools for remote-sensing lithology understanding, rather than autonomous substitutes for professional geological survey.

\section{Declaration of LLM usage}
\label{sec:LLM_Usage}

Large language models and large multimodal models were used in several stages of this work. During dataset construction, GPT-5.4 was used to generate structured geological descriptions and candidate question-answer pairs. Gemini-3-Pro and GPT-5.4 were compared for description generation, and Qwen3-VL-235B-A22B-Instruct was used as a judge model in the blind pairwise comparison. LLM-as-a-Judge modules were also used for quality scoring, candidate filtering, and open-ended answer evaluation.

In addition, LLMs were used to assist with language polishing, grammar correction, and improving the clarity of some manuscript sentences. All key technical designs, data processing procedures, experimental analyses, and final claims were reviewed and verified by the authors. The authors take full responsibility for the content, results, and conclusions of this paper.

\section{Broader impacts}
\label{sec:B_impact}

\lith is designed to support the development and evaluation of geological knowledge-guided multimodal models for remote-sensing lithology interpretation. Its potential positive impacts include promoting benchmark-driven research in geoscientific foundation models, improving the diagnosis of LVLMs in expert-level remote sensing tasks, and supporting applications such as geological survey, regional mapping, mineral exploration, and geoscience education.

At the same time, this benchmark also has potential risks. First, automated lithology interpretation systems may be over-trusted by non-expert users. Remote-sensing-based lithology interpretation is inherently uncertain and should not replace field verification, geological sampling, or expert assessment. Second, models trained or evaluated on \lith may inherit geographic or sensor-specific biases, especially because the current data are mainly based on GF-2 imagery from exposed surface regions. Third, improved geological interpretation tools may be used in resource exploration, which could have environmental and social consequences if not governed properly.

To mitigate these risks, we emphasize that \lith should be used as a research benchmark and decision-support resource rather than an autonomous geological mapping system. Practical deployment should involve domain experts, uncertainty analysis, field validation, and compliance with local environmental and resource-management regulations.

\section{Technical appendices and supplementary material}
\label{sec:technical_appendices}

We provide supplementary materials to improve the transparency and reproducibility of \lith. The supplementary files include additional dataset statistics, representative examples from different capability levels, prompt templates for structured description generation and question-answer construction, quality-scoring rubrics, evaluation scripts, and details of the train-test image-level separation protocol.

The released benchmark follows a unified data format. Each instance contains the image identifier, question type, capability level, lithology category, question, candidate options for multiple-choice questions, reference answer, and metadata used for evaluation. For open-ended questions, we also provide the reference answer and evaluation protocol. For multiple-choice questions, we provide the ground-truth option and standardized answer-parsing rules.

We also provide additional implementation details for model inference and LoRA fine-tuning, including decoding parameters, training hyperparameters, and evaluation settings. These materials are intended to help future researchers reproduce the reported results, conduct fair model comparison, and extend \lith to broader geological remote-sensing scenarios.


\end{document}